\documentclass{article}

\PassOptionsToPackage{numbers, sort&compress}{natbib}
\usepackage[eandd, final]{neurips_2026}

\usepackage[utf8]{inputenc} 
\usepackage[T1]{fontenc}    
\usepackage{hyperref}       
\usepackage{url}            
\usepackage{xurl}           
\usepackage{booktabs}       
\usepackage{amsfonts}       
\usepackage{nicefrac}       
\usepackage{microtype}      
\usepackage{xcolor}         
\usepackage[htt]{hyphenat}
\usepackage{wrapfig}

\title{Formatting Instructions For NeurIPS 2026}

\usepackage{graphicx}
\usepackage{subcaption}

\usepackage{amsmath}
\usepackage{amssymb}
\usepackage{mathtools}
\usepackage{amsthm}
\usepackage{amsfonts}

\usepackage{algorithm}
\usepackage{algorithmic}

\usepackage{multirow}
\usepackage{makecell}
\usepackage{colortbl}
\usepackage{tabularx}
\usepackage{longtable}
\usepackage{listings}
\usepackage{tcolorbox}
\usepackage{enumitem}

\usepackage[capitalize,noabbrev]{cleveref}

\tcbuselibrary{skins,breakable}

\newcommand{\para}[1]{\smallskip\noindent {\bf #1} }

\usepackage{pgfmath}
\definecolor{heatgreen}{HTML}{CDE8CC}
\definecolor{heatmin}{HTML}{F5FAF6}
\newcommand{\heat}[3]{%
  \pgfmathparse{min(max((#1 - #2) / (#3 - #2) * 100, 0), 100)}%
  \pgfmathtruncatemacro{\heatint}{\pgfmathresult}%
  \edef\doheatcolor{\noexpand\cellcolor{heatgreen!\heatint!heatmin}}%
  \doheatcolor#1%
}
\definecolor{applegreen}{rgb}{0.55, 0.71, 0.0}
\definecolor{darkblue}{rgb}{0.0, 0.0, 0.5}
\definecolor{codegreen}{rgb}{0,0.6,0}
\definecolor{codegray}{rgb}{0.5,0.5,0.5}
\definecolor{codepurple}{rgb}{0.58,0,0.82}
\definecolor{backcolour}{rgb}{0.95,0.95,0.92}

\newif\ifshowcomments
\showcommentstrue

\ifshowcomments
\newcommand{\TODO}[1]{{\color{red}{#1}}}
\newcommand{\tianyin}[1]{{\color{cyan!70!blue}{(Tianyin: #1)}}}
\newcommand{\jackson}[1]{{\color{purple}{(Jackson: #1)}}}
\newcommand{\yiming}[1]{{\color{orange}{(Yiming: #1)}}}

\newcommand{\lily}[1]{{\color{violet}{(Lily: #1)}}}
\newcommand{\pial}[1]{{\color{magenta}{(Pial: #1)}}}
\newcommand{\bo}[1]{{\color{red}{(Bogdan: #1)}}}
\newcommand{\yifang}[1]{{\color{purple!70!blue}{(Yifang: #1)}}}
\else
\newcommand{\TODO}[1]{}
\newcommand{\tianyin}[1]{}
\newcommand{\jackson}[1]{}
\newcommand{\yiming}[1]{}
\newcommand{\lily}[1]{}
\newcommand{\pial}[1]{}
\newcommand{\bo}[1]{}
\newcommand{\yifang}[1]{}
\fi

\newcommand{\code}[1]{\texttt{#1}}

\newcommand{\bench}{\textsc{SREGym}}

\newcommand{\symbolimg}[2][0.3cm]{%
  \ensuremath{\vcenter{\hbox{\includegraphics[height=#1]{#2}}}}%
}

\newcommand{\openai}{\symbolimg[0.35cm]{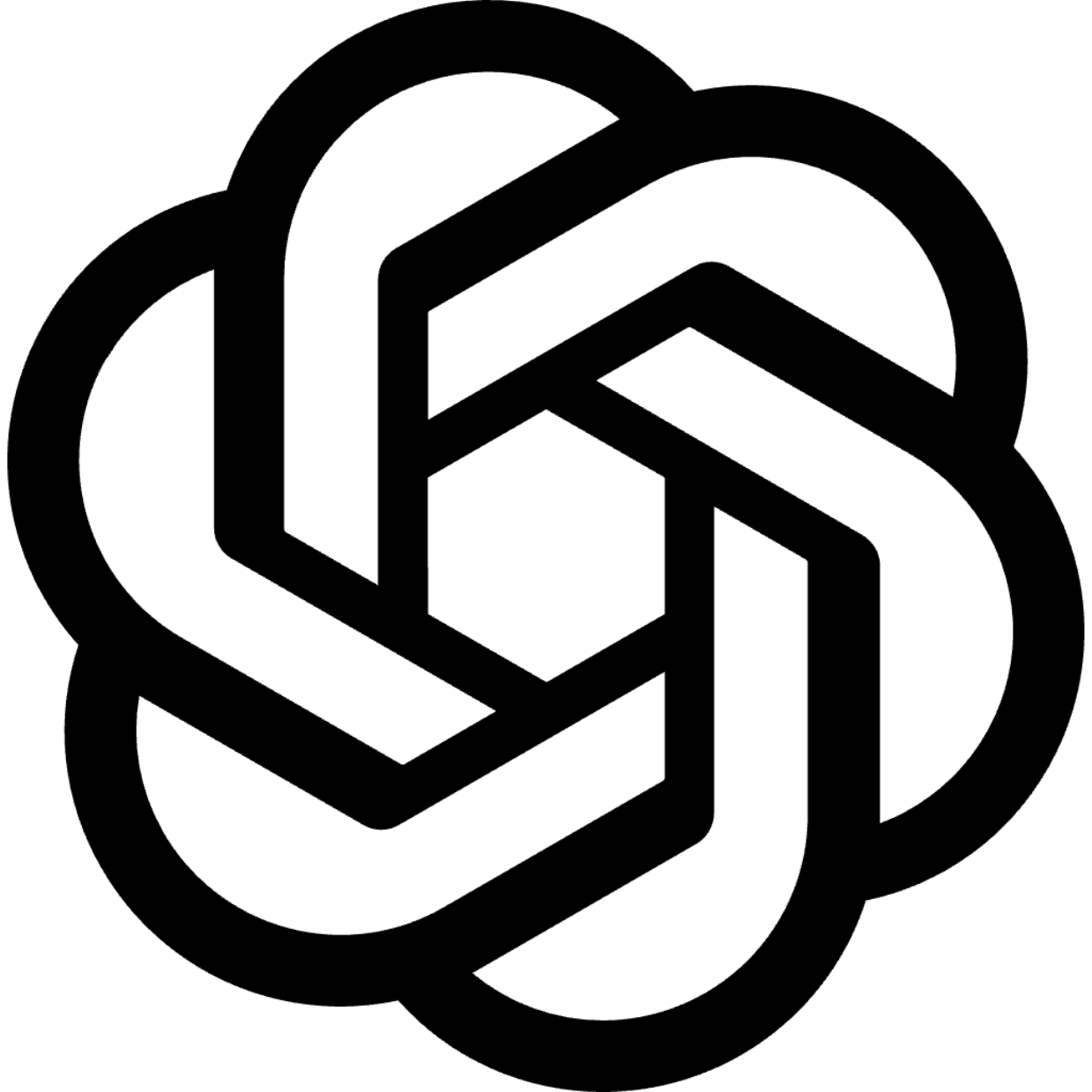}}
\newcommand{\claude}{\symbolimg[0.35cm]{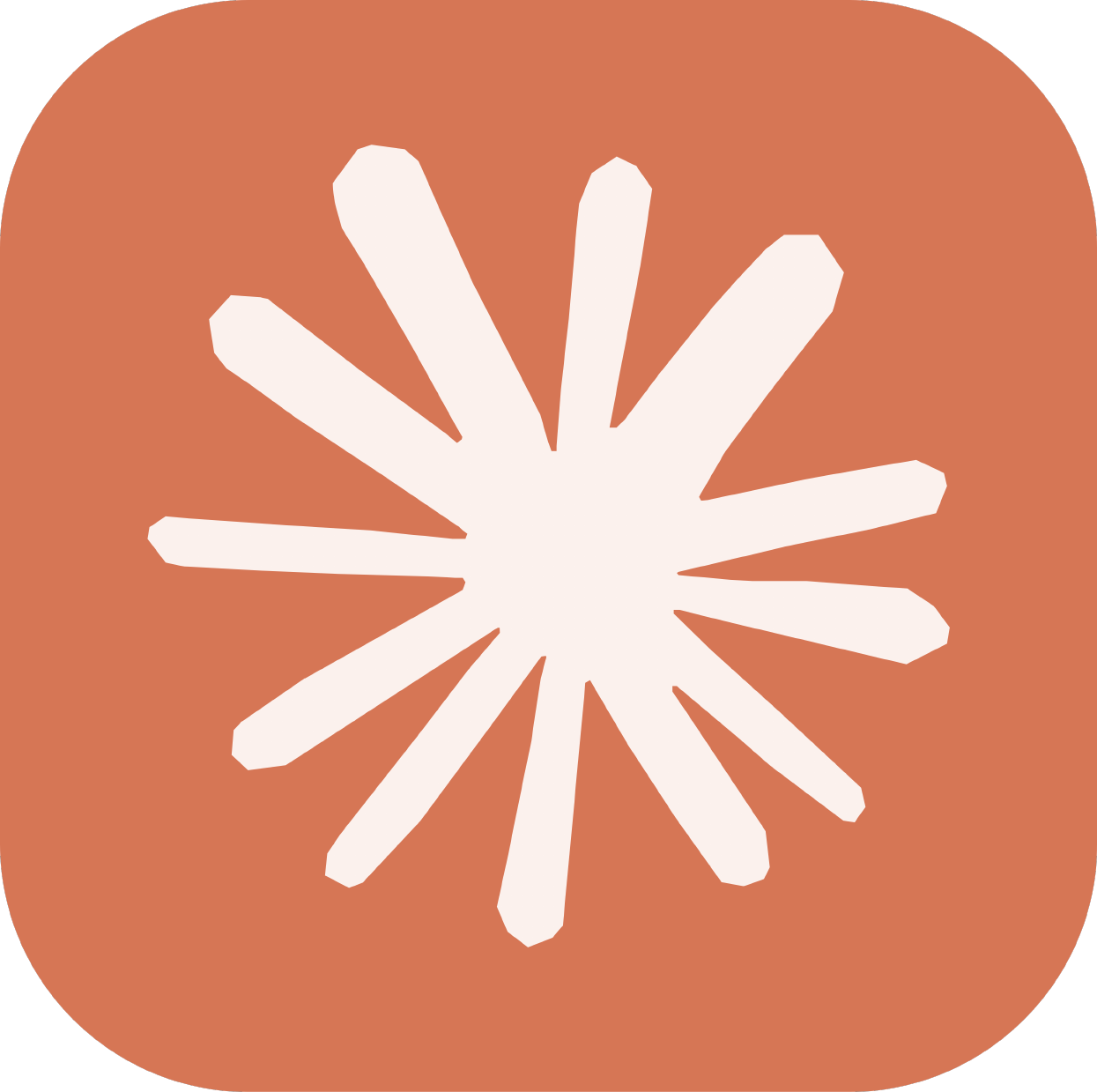}}
\newcommand{\kimi}{\symbolimg[0.35cm]{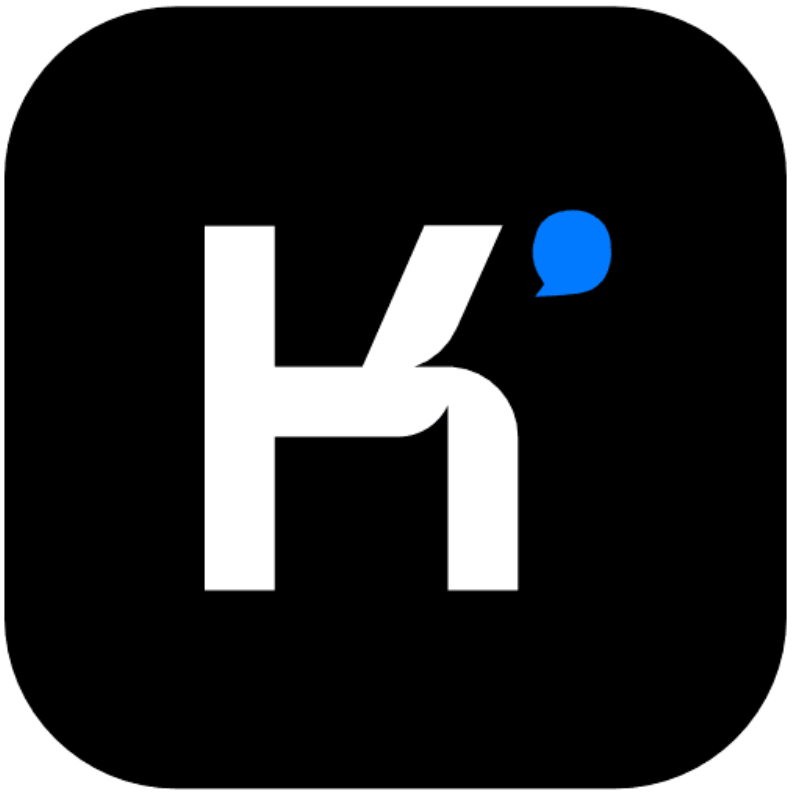}}
\newcommand{\noiseon}{$\blacksquare$}
\newcommand{\noiseoff}{$\square$}

\newtcolorbox{stratussystem}{
    colframe=purple!40!black,colbacktitle=purple!20,
    fontupper=\ttfamily\small,
    coltitle=black,boxrule=1.2pt,enhanced,arc=2pt,fonttitle=\bfseries\large,
    title={\textsc{Stratus} Agent Traces \hfill System Prompt},
    breakable
}

\newtcolorbox{stratushuman}{
    colframe=lime!40!black,colbacktitle=lime!20,
    fontupper=\ttfamily\small,
    coltitle=black,boxrule=1.2pt,enhanced,arc=2pt,fonttitle=\bfseries\large,
    title={\textsc{Stratus} Agent Traces \hfill Human Prompt},
    breakable
}

\newtcolorbox{stratusagent}{
    colframe=red!40!black,colbacktitle=red!20,
    fontupper=\ttfamily\small,
    coltitle=black,boxrule=1.2pt,enhanced,arc=2pt,fonttitle=\bfseries\large,
    title={\textsc{Stratus} Agent Traces \hfill Agent Response},
    breakable
}

\newtcolorbox{stratustool}{
    colframe=blue!40!black,colbacktitle=blue!20,
    fontupper=\ttfamily\small,
    coltitle=black,boxrule=1.2pt,enhanced,arc=2pt,fonttitle=\bfseries\large,
    title={\textsc{Stratus} Agent Traces \hfill Tool Call Message},
    breakable,
}

\theoremstyle{plain}

\theoremstyle{definition}

\theoremstyle{remark}

\usepackage[textsize=tiny]{todonotes}

\title{SREGym: A Live Benchmark for AI SRE Agents\\ with High-Fidelity Failure Scenarios}

\author{%
  Jackson Clark$^{\diamond}$\thanks{Equal contribution.} \and 
  \textbf{Yiming Su$^{\diamond*}$} \and 
  \textbf{Saad Mohammad Rafid Pial$^{\diamond}$} \and  
  \textbf{Yifang Tian$^\dag$} \and 
  \textbf{Lily Gniedziejko$^{\diamond}$} \and       
  \textbf{Hans-Arno Jacobsen$^\dag$} \and    
  \textbf{Yinfang Chen$^{\ddag}$} \and        
  \textbf{Tianyin Xu$^{\diamond}$} \\
  University of Illinois Urbana-Champaign$^{\diamond}$\ \ \ \ \ University of Toronto$^\dag$\ \ \ \ \ Arizona State University$^\ddag$
}
\usepackage{pifont}
\DeclareUnicodeCharacter{2717}{\ding{55}} 
\DeclareUnicodeCharacter{2713}{\ding{51}} 
\usepackage{amssymb}

\begin{document}

\maketitle

\begin{abstract}

AI agents are increasingly used to diagnose and mitigate failures 
    in production systems, known as agentic Site Reliability Engineering (SRE). 
Current SRE benchmarks are limited to oversimplified SRE tasks
    and are unfortunately hard to extend due to bespoke designs.
We present \bench{}, a high-fidelity benchmark
    for SRE agents.
\bench{} exposes a live system environment built atop 
    real-world cloud-native system stacks,
    where high-fidelity failure scenarios are simulated through fault injectors.
\bench{} models the complexity of production environments by simulating 
    (1) a wide range of faults at different layers, 
    (2) various ambient noises,
    and (3) diverse failure modes such as metastable failures and 
    correlated failures.
\bench{} is architected as a modular, extensible framework that orchestrates
    fault and noise injectors across stacks.
\bench{} currently includes 90 realistic, challenging SRE problems.
We use \bench{} to evaluate frontier agents and show
    that their capabilities vary significantly
    in addressing different kinds of failures, with 
    up to 40\% differences in end-to-end results.
\bench{} is actively maintained as an open-source project 
    and has been used by researchers and practitioners. 
\end{abstract}

\section{Introduction}

The software lifecycle extends far beyond writing code,
    encompassing the continuous
    operation of deployed systems in production.
While coding agents 
    have transformed software development~\cite{github-copilot-accenture,stackoverflow-survey-2025},
    accelerated (vibe) coding introduces reliability challenges:
    AI-generated code is reported to introduce
    1.7$\times$ more defects than human-written
    code~\cite{coderabbit-ai-bugs} and
    43\% of code changes made by AI escape testing and cause
    production issues~\cite{lightrun-ai-report}.
Today, major services are already experiencing production outages
    caused by AI-generated code~\cite{amazon-ai-outages}.
The capabilities of agentic AI must extend beyond writing code to
    mitigating production incidents,
    aka 
    Site Reliability Engineering (SRE).

SRE requires capabilities different from coding and Software Engineering (SWE) in general.
SRE agents must reason across multi-modal observability
    data (e.g., system configuration~\cite{xu:osdi:16,sun:osdi:20}, 
        time-series metrics~\cite{Gorilla,jha:sc:20}, unstructured logs~\cite{Sherlog,zhang:2019},
        and distributed traces~\cite{Canopy,magpie,Dapper}),
    interact with domain-specific tools, and execute multi-step mitigation plans whose outcomes
    are only observable at runtime. 
    These requirements make SRE a uniquely challenging domain of agentic AI.


\newlength{\imgboxheight}
\setlength{\imgboxheight}{1.5in}

\begin{figure}[t]
  \centering
  \begin{subfigure}[t]{0.49\textwidth}
    \parbox[c][\imgboxheight][c]{\textwidth}{%
        \centering
        \includegraphics[width=\textwidth]{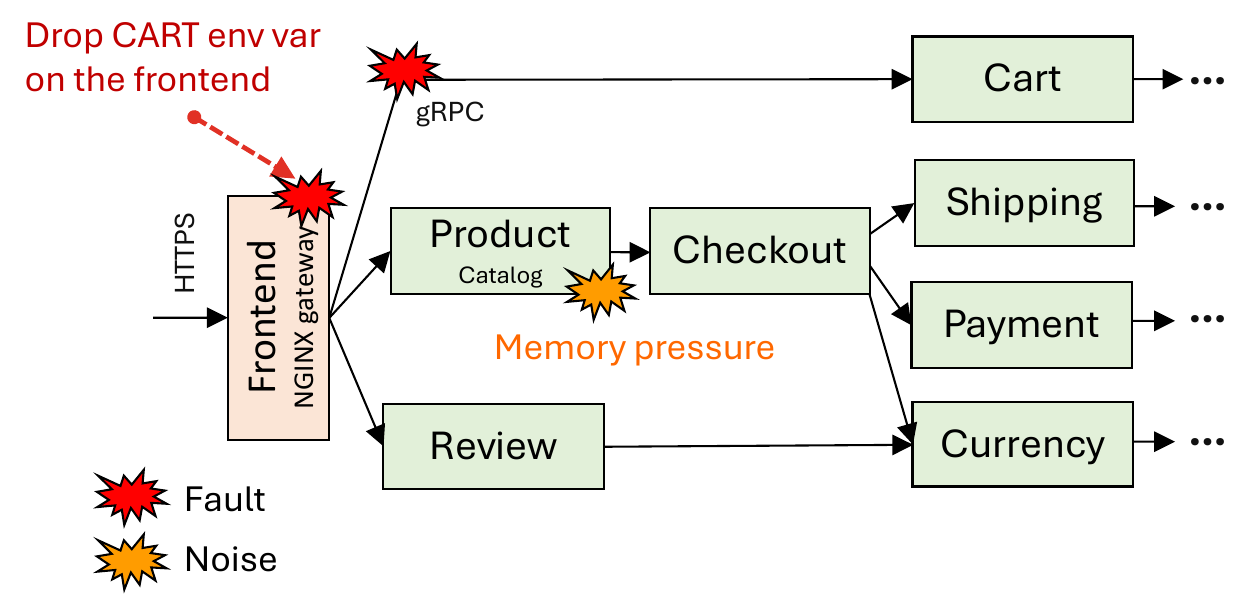}%
    }
    \caption{A failure scenario where
        a microservice is missing an environment variable 
        it needs in order to send requests.}
  \end{subfigure}
  \hfill
  \begin{subfigure}[t]{0.49\textwidth}
    \centering
    \parbox[c][\imgboxheight][c]{\textwidth}{%
        \centering
        \includegraphics[width=\textwidth]{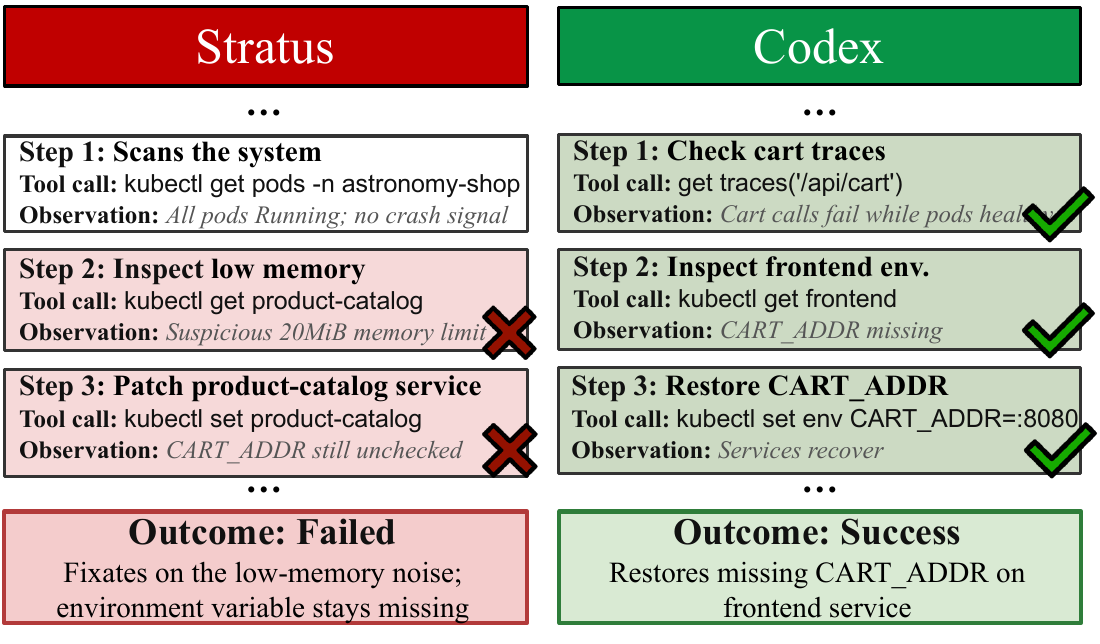}%
    }
    \caption{Agent trajectories of Stratus (with Sonnet-4.6) 
      and Codex (with GPT-5.4) in addressing the problem.}
  \end{subfigure}
  \caption{An SRE problem in \bench{} and the trajectories of two agents when addressing it.}
  \label{fig:intro-trace}
  \vspace{-10pt}
\end{figure}

Existing benchmarks are inadequate and fall behind the development
    of agentic SRE.
Static Q\&A 
    datasets~\cite{liu:arxiv:25,sre-skills-bench} are limited to
    domain knowledge.
Benchmarks for anomaly detection and root cause analysis (RCA)~\cite{han:nips:22,jacob:vldb:21,xu:iclr:25}
    only evaluate whether AI can detect and analyze 
    failures in static datasets, but not whether they can resolve them.
Recent efforts such as AIOpsLab~\cite{chen:mlsys:25}
    and ITBench~\cite{jha:icml:25} take a step forward by
    creating a live system environment with simulated failures.
However, they are limited to {\it oversimplified} SRE tasks.
For example, they primarily focus on application-layer issues,
    whereas real-world failures have diverse root causes across the stack;
    faults in lower layers such as operating systems~\cite{chou:sosp:01} and
    hardware~\cite{hochschild:hotos:21,palix:asplos:11} are known to be harder to address.
Moreover, they mostly inject a single failure into an otherwise
    clean environment, lacking noises and
    unrelated events that coexist in production
    environments~\cite{gunawi:socc:16,liu:hotos:19,huang:hotos:17}.
Unfortunately, these benchmarks are 
    hard to extend due to their bespoke designs, e.g., 
    hardcoding failure simulation in problem-specific scripts
    and lacking support for distributed event coordination.

In this paper, we present \bench{}, a high-fidelity 
    benchmark for SRE agents.
\bench{} shares the high-level principle of prior work~\cite{chen:mlsys:25,jha:icml:25} 
    to expose a {\it live} system environment 
    built atop real-world cloud-native system stacks, where failures are simulated through fault 
    injectors.\footnote{We follow the terminology of the classic Fault-Error-Failure model~\cite{aviz:04,laprie:95}
        where {\it faults} are root causes
        (e.g., software bugs, hardware malfunctions, and misconfigurations); 
        a fault can cause abnormal behaviors referred to as {\it errors}
        which (if not handled properly) propagate further and, once visible to users, are referred to
        as {\it failures}.}
Differently, 
    \bench{} models the complexity of dynamic, noisy, and eventful 
    production environments to achieve {\it high-fidelity}
    failure scenarios that not only challenge AI but also ensure the relevance of the problems. 
\bench{} currently includes 90 realistic, challenging SRE problems.
Figure~\ref{fig:intro-trace} shows one problem and the corresponding agent behavior.
\bench{} presents three new features:

\begin{enumerate}[leftmargin=*,topsep=-2pt,itemsep=1pt]
    \item {\bf Simulating a wide range of faults across the system stack}, including hardware faults~\cite{hochschild:hotos:21},
        OS kernel faults~\cite{chou:sosp:01,palix:asplos:11}, and misoperations~\cite{gu:sosp:23,sun:osdi:22}, 
        in addition to application issues.
    \item {\bf Simulating various ambient noises}, low-impact faults that are unrelated to the root causes of target failures,
        which introduce ambiguity and potentially cause distractions.
    \item {\bf Supporting diverse failure modes} such as metastable behavior~\cite{bronson:hotos:21,huang:osdi:22,Isaacs:hotos:25} 
        and concurrent, correlated failures~\cite{zhai:nsdi:20,ford:osdi:10} by orchestrating distributed events (e.g., faults and noises).
\end{enumerate}


\bench{} is architected as a modular, extensible
framework that composes fault and noise injectors across stacks
    into high-fidelity failure scenarios as SRE problems.
The modularity and extensibility are not only matters of engineering discipline for usability and maintainability
    (which are critical to \bench{} as a community-driven benchmark; see \S\ref{sec:engineering-practices}),
    but also key enablers of its mission as a useful benchmark.
For instance, noises must be composed alongside target failures
    and the manifestation of metastable behavior requires 
    temporal coordination of multiple correlated faults and events.
\bench{} provides a unified programming interface to 
    curate high-quality SRE problems by mutating existing failure scenarios (e.g., by altering noises)
    and creating new ones.


We use \bench{} to evaluate an SRE agent (Stratus~\cite{chen:nips:25})
    and two coding agents (Claude Code~\cite{claude-code} and OpenAI Codex~\cite{gpt-codex}),
    with different models (including Sonnet-4.6, GPT-5.4, and Kimi K2.5).
The success rates of diagnosis and mitigation range from
    38.1\%--72.6\% and 40.4\%--78.5\%, respectively, across agent-model pairs.
The agents show strong abilities in addressing
    application issues,
    but their performance drops significantly
    on failures rooted in other layers or failure modes, with 
    up to 40\% differences in end-to-end results. 
For compound failures, agents tend to draw partial conclusions, missing 
    opportunities to address them comprehensively.
Similarly, agents are affected by noises in nontrivial ways.
Overall, by challenging frontier AI agents and models with high-fidelity failure scenarios,
    \bench{} provides a foundation for advancing agentic SRE technologies
    toward production readiness.


\bench{} is actively maintained as an open-source project at 
    \url{https://github.com/SREGym/SREGym}.
It has been used by researchers and practitioners,
    including in the releases of Devin Oncall~\cite{devin-oncall-blog} and Datadog's Scout~1.0~\cite{datadog-scout}.

\section{\bench{}}
\label{sec:platform}

Building a high-fidelity SRE benchmark is challenging.
The benchmark must create a realistic operational environment
    and simulate sophisticated failure scenarios.
We enforce the following design principles in developing \bench{} (\S\ref{sec:engineering-practices} discusses
    the engineering practices behind them):
\begin{itemize}[leftmargin=*,topsep=0pt,itemsep=1pt]
    \item {\bf Creating noisy and eventful environments.} A clean, quiet environment is not only unrealistic
        but also insufficiently challenging for AI. However, few existing benchmarks model noises.
        \bench{} offers controlled noises without compromising reliable evaluation.
    \item {\bf Simulating faults, not symptoms.} Unlike benchmarks that use chaos engineering tools
        to create symptoms that can only be mitigated by stopping the tools,
        \bench{} simulates fine-grained faults.
    \item {\bf Composability is key to scaling problems.}
        \bench{} achieves composability with support for orchestrating faults and noises into different failure scenarios.
        Prior work uses Ansible scripts, which are hard to extend
        and cannot easily support failure modes that require distributed events.
    \item {\bf Usability and extensibility are essential.} Usability/extensibility may not reflect academic novelty
        but are critical to the success of a benchmark.
        \bench{} is push-button and offers APIs for problem extension (which enables
        its users to contribute new problems).
\end{itemize}

\begin{figure}[H]
    \centering
    \includegraphics[width=\columnwidth]{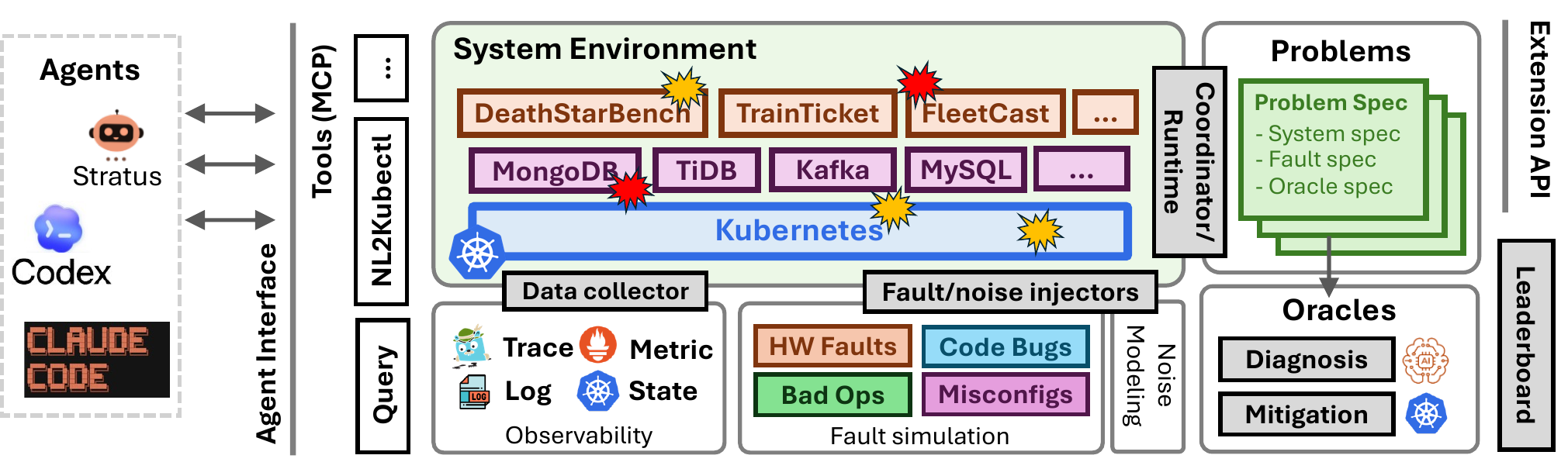}
    \vspace{-7.5pt}
    \caption{Overview of the \bench{} framework and benchmark suites.}
    \label{fig:overview}
    \vspace{-2.5pt}
\end{figure}

Figure~\ref{fig:overview} provides an overview of \bench{} in terms of its components.
We present the main components in the remainder of this section.

\subsection{Problem Definition: A User Perspective}
\label{sec:problem-definition}

\bench{} currently contains 90 SRE problems.
A problem in \bench{} creates a failure scenario
    in a live production-like environment
    that an SRE agent must diagnose and mitigate.
Formally, a problem is a four-tuple
    $P = (\mathcal{E}, \mathcal{I}, \mathcal{F}, \mathcal{O})$,
    including:
\begin{itemize}[leftmargin=*,topsep=-2pt,itemsep=1pt]
\item {\bf System environment $\mathcal{E}$.} \bench{}
    exposes a production-like environment that deploys
    user applications, backend systems,
    and control/management services (e.g., Kubernetes controllers).


\item {\bf Agent interface $\mathcal{I}$.}
The exposed tools and APIs for agents to observe and interact
    with $\mathcal{E}$.

\item {\bf Faults and noises $\mathcal{F}$.}
A set $\mathcal{F} = \{f_1, \ldots, f_k\}$ of one or more
    faults, each targeting a specific component in $\mathcal{E}$.
A subset of $\mathcal{F}$ may be designated as {\it noises}:
    transient, low-impact disturbances that an agent must
    distinguish from the root cause(s) of the target failure.

\item {\bf Oracles $\mathcal{O} = (\mathcal{O}_d, \mathcal{O}_m)$},
    where $\mathcal{O}_d$ is a diagnosis oracle and $\mathcal{O}_m$
    is a mitigation oracle.
\end{itemize}

For each problem, the inputs and expected outputs of evaluated agents are
    described as follows.
\begin{itemize}[leftmargin=*,topsep=-2pt,itemsep=1pt]
\item {\bf Inputs.} The agent can query the system environment $\mathcal{E}$ (with $\mathcal{F}$)
    through standard observability modalities via the agent interface $\mathcal{I}$.

\item {\bf Expected outputs.}
The agent makes two submissions for a problem.
The first is a natural-language diagnosis describing the root cause
    of the failure.
This submission triggers $\mathcal{O}_d$ (see \S\ref{sec:oracle}).
The second submission signals that the agent has completed
    its mitigation effort
    and triggers $\mathcal{O}_m$ based on actual system and application states.
Following best practices for agent evaluation~\cite{anthropic-evals,abc-checklist},
    \bench{} uses programmatic verification whenever possible for reliable evaluation.
\end{itemize}

\subsection{System Environments}
\label{sec:implementation}

\bench{} exposes a cloud-native, Kubernetes-based system environment,
    where applications are deployed in Docker containers.
An SRE problem selects which application(s) to deploy
    based on the failure scenarios (\S\ref{sec:problems}).
\bench{} ships a catalog of cloud-native applications,
    including
    DeathStarBench~\cite{gan:asplos:19},
    Train Ticket~\cite{trainticket},
    Astronomy Shop~\cite{otel-demo},
    and in-house applications (a satellite orbit
    simulator and a flight booking service),
    paired with backend data systems
    (e.g., MongoDB, TiDB, Kafka, and MySQL).
Each application has corresponding workloads (user traffic).
The applications and backend systems are managed by Kubernetes
    operators.
The applications, systems, and operators are deployed using the
    Helm package manager~\cite{helm}.
Built atop the {\it de facto}
    cloud-native stack, the environment is highly extensible---it supports any real-world cloud-native
    applications/systems with Kubernetes manifests or Helm charts,
    and can be deployed on any mainstream hardware platform.

\subsection{Agent Interface}
\label{sec:agent-interface}

\bench{} makes no assumptions about the architectures or interaction patterns of evaluated agents,
    to avoid brittle coupling of agent and benchmark.\footnote{For example, AIOpsLab~\cite{chen:mlsys:25} decomposes
        each failure scenario into four isolated sub-problems (detection, localization, RCA, and mitigation)
        and scores them independently, but real-world failure handling is a single end-to-end loop in which
        earlier evidence and actions shape later decisions; \bench{} instead evaluates the entire failure scenario
        holistically.}
The benchmark exposes Model Context Protocol (MCP) servers for agents,
    allowing them to observe and interact with the target systems and their environment.
\bench{} provides the following interfaces as MCP servers:
\begin{itemize}[leftmargin=*,topsep=-2pt,itemsep=1pt]
    \item {\bf Metrics} for querying time-series performance metrics through Prometheus~\cite{prometheus}.
    \item {\bf Logs} for searching and filtering container logs through Loki~\cite{loki}.
    \item {\bf Traces} for inspecting request traces between system components via Jaeger~\cite{jaeger}.
    \item {\bf Cluster control} for executing arbitrary commands to
        observe and change system state. We currently support \texttt{kubectl} commands
        (Kubernetes' command-line interface).
    \item {\bf Submission} for submitting diagnosis and mitigation results,
        which triggers evaluation.
\end{itemize}

For power users who want to design customized tools,
    \bench{} also exposes the raw observability endpoints
    and Kubernetes API endpoints for their agents to connect to.

\begin{figure}[t]
\centering
\begin{minipage}[t]{0.49\columnwidth}
\vspace{0pt}
    \centering
\definecolor{vsKeyword}{HTML}{CC0000}    
\definecolor{vsString}{HTML}{A31515}     
\definecolor{vsComment}{HTML}{008000}    
\definecolor{vsDecorator}{HTML}{AF00DB}  
\definecolor{vsSelf}{HTML}{0070C1}       
\definecolor{vsClassName}{HTML}{267F99}  
\definecolor{vsDefault}{HTML}{000000}    
\definecolor{appBoxColor}{HTML}{0097A7}
\definecolor{oracleBoxColor}{HTML}{2E7D32}
\definecolor{faultBoxColor}{HTML}{7B1FA2}

\lstdefinestyle{vscodelight}{
    language=Python,
    basicstyle=\ttfamily\footnotesize\color{vsDefault},
    keywordstyle=\bfseries\color{vsKeyword},
    keywordstyle=[2]\bfseries\color{vsSelf},
    keywordstyle=[3]\color{vsClassName},
    commentstyle=\color{vsComment},
    stringstyle=\color{vsString},
    showstringspaces=false,
    breaklines=true,
    columns=fullflexible,
    keepspaces=true,
    tabsize=4,
    frame=none,
    xleftmargin=0pt,
    xrightmargin=0pt,
    aboveskip=0pt,
    belowskip=0pt,
    lineskip=-2pt,
    morekeywords={class,def,super,print},
    morekeywords=[2]{self},
    morekeywords=[3]{K8STargetPortMisconfig,Problem,SocialNetwork,LLMAsAJudgeOracle,TargetPortMisconfigMitigationOracle,VirtualizationFaultInjector,K8sNetworkPortMisconfig},
    moredelim=[l][\bfseries\color{vsDecorator}]{@mark},
    escapeinside={!*}{*!},
}

\begin{tikzpicture}[remember picture]
\node[inner sep=0pt, outer sep=0pt, anchor=north west] (code) {%
\begin{minipage}{\linewidth}
\begin{lstlisting}[style=vscodelight]
class K8sNetworkPortMisconfig(Problem):
  def __init__(self):!*\tikz[remember picture] \coordinate (app-start);*!
    self.app = SocialNetwork()
    self.app.create_workload()!*\tikz[remember picture] \coordinate (app-end);*!
!*\tikz[remember picture] \coordinate (ora-start);*!
    self.root_cause = ("The user-service has a misconfigured network port [...]")
    self.diagnosis_oracle = LLMAsAJudgeOracle(problem=self, expected=self.root_cause)
    self.mitigation_oracle = MitigationOracle(problem=self)!*\tikz[remember picture] \coordinate(ora-end);*!
   
  @mark_fault_injected
  def inject_fault(self):!*\tikz[remember picture] \coordinate (fault-start);*!
    injector = NetworkPortFaultInjector(namespace=self.namespace)
    injector._inject(
      fault_type="port-misconfig", 
      microservice="user-service")!*\tikz[remember picture] \coordinate (fault-end);*!
\end{lstlisting}
\end{minipage}%
};
\end{tikzpicture}%
\begin{tikzpicture}[remember picture, overlay]
\draw[appBoxColor, thick, rounded corners=1pt]
  ([xshift=-79pt, yshift=-2pt]app-start) rectangle
  ([xshift=2pt, yshift=-2pt]code.east |- app-end);
\node[anchor=south east, fill=appBoxColor, inner sep=1.5pt,
  font=\sffamily\fontsize{7}{6}\selectfont\bfseries\color{white}]
  at ([xshift=2pt, yshift=-2pt]code.east |- app-start) {Application};
\draw[oracleBoxColor, thick, rounded corners=1pt]
  ([xshift=10pt, yshift=2pt]ora-start) rectangle
  ([xshift=2pt, yshift=-2pt]code.east |- ora-end);
\node[anchor=south east, fill=oracleBoxColor, inner sep=1.5pt,
  font=\sffamily\fontsize{7}{6}\selectfont\bfseries\color{white}]
  at ([xshift=2pt, yshift=2pt]code.east |- ora-start) {Oracle};
\draw[faultBoxColor, thick, rounded corners=1pt]
  ([xshift=-97pt, yshift=-2pt]fault-start) rectangle
  ([xshift=2pt, yshift=-2pt]code.east |- fault-end);
\node[anchor=south east, fill=faultBoxColor, inner sep=1.5pt,
  font=\sffamily\fontsize{7}{6}\selectfont\bfseries\color{white}]
  at ([xshift=2pt, yshift=-2pt]code.east |- fault-start) {Fault};
\end{tikzpicture}
\vspace{-10pt}
\captionof{figure}{Implementing a problem in \bench{}
    (noises are injected by the framework).}
\label{fig:example-problem}
\end{minipage}%
\hfill
\begin{minipage}[t]{0.479\columnwidth}
    \vspace{0pt}
    \centering
    \captionof{table}{Fault and noise injectors in \bench{}
    (``K8s'' refers to Kubernetes).}
    \label{tab:fault-mechanism}
    \footnotesize
    \renewcommand{\arraystretch}{1.1}
\setlength{\tabcolsep}{4pt}
\begin{tabular}{ll}
\toprule
\textbf{Mechanism} & \textbf{Simulated Faults} \\
\midrule
Kill a process or a pod & Fail-stop behavior\\
Stress hardware~\cite{stress-ng}  & Fail-slow behavior~\cite{210508} \\
Fail syscall via eBPF             & OS/hardware faults~\cite{chou:sosp:01,palix:asplos:11}\\
Corrupt a disk sector~\cite{dm-dust} & Disk sector errors~\cite{Schroeder:2010,Bairavasundaram:2007} \\
Fault in deploy.yaml              & Service mis-deployment \\
Fault in app. config & App. misconfiguration~\cite{xu:sosp:13} \\
Fault in K8s config & K8s misconfiguration~\cite{k8s-config-bug} \\
Use buggy app. code & Code bugs \\
Use buggy app. operator & Misoperations~\cite{gu:sosp:23,gu:nsdi:26} \\
Increase client loads & Service overloads~\cite{meza:osdi:23} \\
\midrule
Pause/restart unrelated pods & Transient pod failures \\
Inject latency / drop packets & Network delay / jitter \\
Stress node resources & Noisy neighbors~\cite{Heracles,CLITE} \\
\bottomrule
\end{tabular}
\end{minipage}
\vspace{-10pt}
\end{figure}

\subsection{Creating SRE Problems by Composing Faults}
\label{sec:problems}

\bench{} offers 90 problems (with new ones being continuously added).
The problems can be easily mutated, e.g., with different noise patterns.
\bench{} currently provides 47 fault primitives that
    can be applied to 139 deployable services across 5 supported
    applications.
With compatibility constraints (certain faults can only be applied to specific services),
    \bench{} offers 3{,}623 viable fault-component pairs even before composing noises
    or multiple faults, roughly 40$\times$ the 90 curated problems (see Appendix~\ref{sec:appendix:coverage}).
This is a structural difference from benchmarks that ship as a
    fixed set of hardcoded scenarios:
    the curated 90 problems reflect what we have validated
    end-to-end, not the extent of what \bench{} can express.
Figure~\ref{fig:example-problem} shows the implementation of one \bench{} problem.


{\bf Fault and Noise Simulation.}
Table~\ref{tab:fault-mechanism} shows the faults and noises \bench{} simulates
    and the simulation mechanisms.
The faults include
    common ones like application misconfigurations
    and new ones \bench{} implements (e.g., a tool that uses
    eBPF to simulate OS and hardware faults).
More fault injectors can be directly integrated into \bench{}.
These faults are at different layers in the system stack
    and manifest via various symptoms.
They require SRE agents to have comprehensive
    knowledge and understanding of different system components and their interactions.
For example, faults injected into microservices require SRE agents
    to understand application logic and its dependencies~\cite{Treynor:2017,veeraraghavan:osdi:18}.
Faults injected into Kubernetes controllers and operators
    require SRE agents to understand
    how Kubernetes manages the cluster and applications~\cite{gu:nsdi:26,sun:osdi:22}.
Lower-level OS and hardware faults require
    SRE agents to understand vertical interactions of applications, OSes, and hardware~\cite{gunawi:08}.
Each injector exposes a Python API;
    composing a compound failure reduces to
    invoking several injectors.

We define noises as transient, self-recovering disturbances (e.g., a pod that crashes and is
    rescheduled, or a few dropped requests)---they are not failure root causes.
\bench{} injects these transient events with a configurable schedule.
The agents see symptoms from both offending faults and
    noises and must distinguish the two types.

\begin{figure}
    \centering
    \begin{subfigure}[t]{0.49\textwidth}
        \centering
        \includegraphics[width=\textwidth]{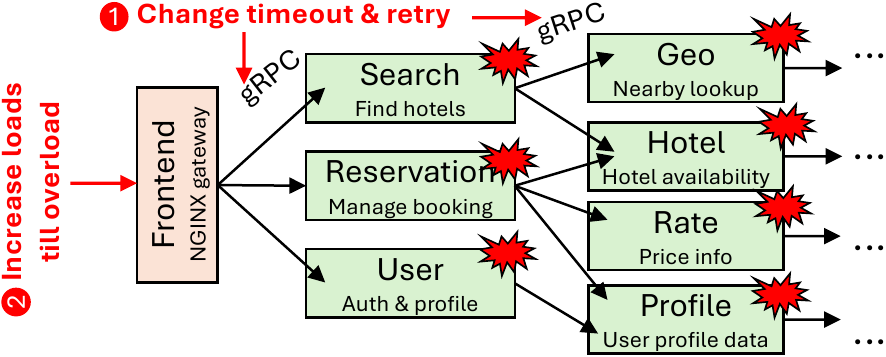}
        \caption{Metastable failures}
        \label{fig:metastable-fault}
    \end{subfigure}
    \hfill%
    \begin{subfigure}[t]{0.49\textwidth}
        \centering
        \includegraphics[width=\textwidth]{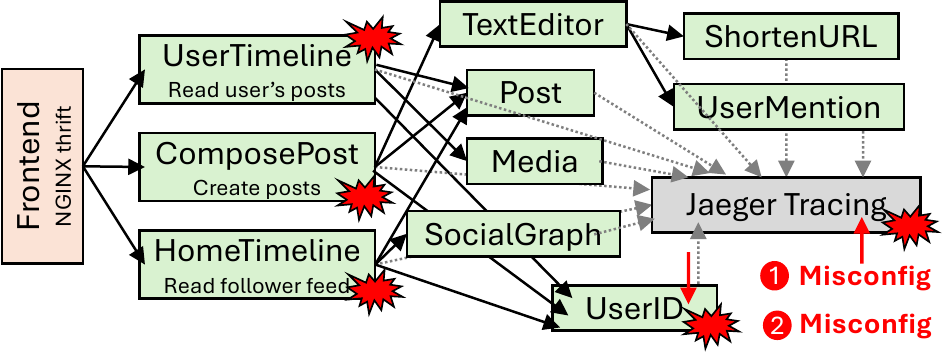}
        \caption{Concurrent failures}
        \label{fig:noisy-environment}
    \end{subfigure}
    \caption{SRE problems that create scenarios with different failure modes in \bench{}.}
    \label{fig:failure-modes}
    \vspace{-15pt}
\end{figure}

{\bf Failure Modes.}
\bench{} enables developers to compose fault and noise injectors
    to create failure scenarios in different modes.
Our current problem set covers three important failure modes.
\begin{itemize}[leftmargin=*,topsep=-2pt,itemsep=1pt]
    \item {\bf Metastable failures} are self-sustaining congestive collapses
        in which the system degrades in response to transient events (e.g., a load surge)
        but fails to recover after the trigger is removed~\cite{bronson:hotos:21,huang:osdi:22,Isaacs:hotos:25}.
        Metastable failures are known to be hard to diagnose,
            as they do not manifest as crashes.
            Figure~\ref{fig:metastable-fault} shows a metastable failure problem we created,
                which (1) sets overly aggressive gRPC configurations
                for connection timeout (50ms) and retry count (30),
                (2) pushes the application (Hotel Reservation in DeathStarBench~\cite{gan:asplos:19})
                into a vulnerable state with a high load of 3000 requests per second,
                and (3) triggers the metastable failure with transient CPU stress.
            In this way, all RPC requests time out and are retried at the same time, causing a retry storm.
            \bench{}'s ability to coordinate different events is crucial to creating
    this problem.
    \item {\bf Concurrent failures} are compound failures caused
        by multiple independent failures occurring simultaneously.
    Figure~\ref{fig:noisy-environment} depicts a problem in which two faults are injected
    into the application (Social Network in DeathStarBench~\cite{gan:asplos:19}):
    (1) a scheduler misconfiguration that makes an observability service
        pod unschedulable, and
    (2) a network port misconfiguration that causes the \texttt{UserID} service to fail, which in turn
        fails user requests.
    The problem evaluates whether SRE agents can understand and
        prioritize the port misconfiguration
        as it directly affects service availability (the scheduler
        misconfiguration only affects the observability service, which is invisible to end users).
    \item {\bf Correlated failures} are failures in which multiple components
        fail at the same time because they share a common cause or dependency~\cite{zhai:nsdi:20,ford:osdi:10}.
        The agent must recognize the correlations between the symptoms and
        connect them to the underlying root cause(s).
\end{itemize}

\subsection{Evaluation Oracles}
\label{sec:oracle}

Designing oracles that fairly evaluate SRE agents is challenging.
Existing SRE benchmarks use oracles that evaluate diagnosis results
    by requiring SRE agents to strictly match
    text labels (e.g., names of offending services and root causes)~\cite{chen:mlsys:25,cloudopsbench};
    such oracles are brittle because predefined labels are
    often ambiguous and not mutually exclusive;
    our experience shows that agents often report correct results,
    but are scored as incorrect due to not matching labels exactly.
ITBench~\cite{jha:icml:25} proposed Normalized Topology-Aware
    Matching, which requires fine-grained annotations of
    failure-propagation graphs.
However, such laborious annotation is error-prone and hard to scale.





{\bf Diagnosis Oracle.}
\bench{} adopts a checklist-based
    LLM-as-a-judge protocol~\cite{zheng:nips:23}
    that decomposes diagnosis evaluation into
    fine-grained questions in a structured form, and produces stable verdicts
    across evaluators.
We follow the principles of
    decomposing evaluation into multi-dimensional questions to improve inter-evaluator 
    agreement~\cite{lee:emnlp:25}, and grounding the rubric in
    domain-expert insight instead of LLM-generated criteria~\cite{chu:chi:25,tian:arxiv:25}. 
Given a ground-truth root-cause description~$g$
    and an agent-submitted diagnosis~$d$,
    the diagnosis oracle assembles a prompt containing~$g$, $d$,
    and a checklist of~$N=9$ Yes/No questions
    organized into~$K=3$ dimensions.
An LLM evaluator returns, for each question~$q$,
    a judgment~$y_q\!\in\!\{0,1\}$
    together with supporting evidence and a confidence score.
For dimension $k$ with a set of questions $Q_k$, 
    the per-dimension score $s_k$ is the fraction of affirmative answers:
    $s_k = \frac{1}{|Q_k|}\sum_{q\in Q_k} y_q$,
%
and the {\it aggregated score} is a weighted sum
%
    $S = \sum_{k=1}^{K} w_k\, s_k$ (we give all dimensions equal weight, i.e., $w_k=\frac{1}{3}$).
%
The final verdict is $\hat{v}=\mathbb{1}[S \ge \tau]$
    with default threshold~$\tau\!=\!\frac{7}{9}$ (which prevents 
    a submission from passing while missing an entire dimension).

We find that decomposition into questions
    yields transparent, per-dimension evaluation
    of \emph{where} the agent's understanding fell short.
Our three dimensions are:
%
\begin{itemize}[leftmargin=*,topsep=-2pt,itemsep=0pt]
    \item \textbf{Fault localization}:
        Does the diagnosis identify the correct
        originating component, distinguishing it from
        downstream victims of cascading failures?
    \item \textbf{Fault characterization}:
        Does the diagnosis capture the faults
        and concrete details (e.g., the wrong port value,
        a specific environment variable)?
    \item \textbf{Failure scope}:
        Does the diagnosis avoid over- or under-attributing
        impact in terms of the components and symptoms?
\end{itemize}
%



\begin{wraptable}{r}{6.5cm}
\vspace{-20pt}
\centering
\small
\caption{Pairwise inter-evaluator agreement.}
\vspace{-5pt}
\label{tab:judge-reliability}
\begin{tabular}{llrr}
\toprule
Judge A & Judge B & Agree & $\kappa$ \\
\midrule
\claude{} Sonnet-4.6 & Human             & \textbf{0.95} & \textbf{0.90} \\
\claude{} Sonnet-4.6 & \openai{} GPT-5.4 & 0.88          & 0.76          \\
\claude{} Sonnet-4.6 & \kimi{} K2.5      & 0.91          & 0.82          \\
\midrule
\openai{} GPT-5.4    & \kimi{} K2.5      & \textbf{0.97} & \textbf{0.94} \\
\openai{} GPT-5.4    & Human             & 0.85          & 0.70          \\
\kimi{} K2.5         & Human             & 0.88          & 0.76          \\
\bottomrule
\end{tabular}
\end{wraptable}

\label{sec:eval-judge-validation}%
Table~\ref{tab:judge-reliability} validates the oracle's verdicts on a
    stratified sample of $100$ agent diagnosis results, 
    independently labeled by a domain expert and scored by two alternate
    LLM evaluators (GPT-5.4 and Kimi K2.5). 
The default oracle agrees with human experts at Cohen's
    $\kappa\!=\!0.90$ (almost perfect agreement), 
    while the two alternate LLM evaluators converge at
    $\kappa\!=\!0.94$ and retain substantial agreement
    with human experts ($\kappa\!=\!0.70$ and $0.76$ for
    GPT-5.4 and Kimi K2.5, respectively).
The convergence between LLM evaluators suggests that 
    the decision depends primarily on checklist decomposition, not the 
    choice of LLM (a concern with
    single-model LLM-as-a-judge evaluation~\cite{zheng:nips:23}).

{\bf Mitigation Oracle.}
The mitigation oracle is problem-specific to accurately
    reflect whether the target failure is truly mitigated.
The oracle checks whether the target fault is resolved and
    whether the target system has recovered to a healthy state.
The mitigation oracle uses both client-side observability such as
    user request success rate
    and system-side observability of application processes,
    and the Kubernetes cluster.
This allows us to evaluate the agent's mitigation results on complicated
    failure scenarios, such as metastable failures and low-level software/hardware failures.

\section{Results}
\label{sec:eval-setup}

We use \bench{} to evaluate three AI agents powered by frontier LLMs:
Stratus~\cite{chen:nips:25}, a state-of-the-art SRE agent
    with two LLMs: Claude Sonnet-4.6
    and Kimi K2.5;
    and two coding agents:
    Claude Code~\cite{claude-code} with Claude Sonnet-4.6
    and Codex~\cite{gpt-codex} with GPT-5.4 (gpt-5.4-2026-03-05).
Each agent-model pair is evaluated with three runs per problem.
We use Claude Sonnet-4.6 
    for the diagnosis oracle (see \S\ref{sec:oracle}) consistently across the evaluation.

For noise simulation, we randomly inject two noise patterns 
    every five minutes, each lasting two minutes. 
The noise simulation is done by the framework, 
    not hardcoded in any problem. 

{\bf Evaluation Metrics.}
We evaluate the agents on the success rates of
    diagnosis and mitigation tasks.
Diagnosis success rate measures whether the agent pinpoints the
    root causes of target failures correctly.
Mitigation success rate measures whether the agent successfully
    mitigates failures (verified by the mitigation oracle).
We also measure end-to-end (E2E) success rate, referring to
    cases where the agent achieves both correct diagnosis
    and correct mitigation on the same run.
We also report Time-To-Diagnose (TTD),
    Time-To-Mitigate (TTM),
    and mean token usage per problem run.


\subsection{Overall Benchmarking Results}
\label{sec:overall-results}

\begin{table*}[t]
    \caption{Overall benchmark results on~\bench{}.
    TTD and TTM are capped at the 1800-second agent timeout,
    with timed-out runs contributing the cap value,
    so failure cost is reflected in the mean rather than inducing survivorship bias.
    \noiseon : runs with noise injected;
    \noiseoff : runs without noise injected.}
    \label{tab:main-results}
    \centering
    \footnotesize
    \setlength{\tabcolsep}{3pt}
    \begin{tabular}{llccccccc}
        \toprule
        Agent & Model & Noise & Diag. (\%)\ $\uparrow$ & Mitig. (\%)\ $\uparrow$ & E2E (\%) $\uparrow$ & TTD (s) $\downarrow$ & TTM (s) $\downarrow$ & \# Tokens $\downarrow$ \\
        \midrule
        \multirow{4}{*}{Stratus} & \multirow{2}{*}{\claude{} Sonnet-4.6} & \noiseoff & \heat{61.5}{38.1}{72.6}\% & \textbf{\heat{78.5}{40.4}{78.5}\%} & \heat{54.8}{26.7}{60.7}\% & \textbf{114.0} & 771.1 & 812K \\
                                 &                                    & \noiseon  & \heat{51.5}{38.1}{72.6}\% & \heat{61.1}{40.4}{78.5}\% & \heat{39.6}{26.7}{60.7}\% & 170.5 & 885.0 & 464K \\
                                 \cmidrule{2-9}
                                 & \multirow{2}{*}{\kimi{} K2.5} & \noiseoff & \heat{40.4}{38.1}{72.6}\% & \heat{40.4}{40.4}{78.5}\% & \heat{27.4}{26.7}{60.7}\% & 674.5 & 1348.8 & \textbf{413K} \\
                                 &                            & \noiseon  & \heat{38.1}{38.1}{72.6}\% & \heat{41.9}{40.4}{78.5}\% & \heat{26.7}{26.7}{60.7}\% & 656.4 & 1283.2 & 443K \\
        \midrule
        \multirow{2}{*}{Claude Code} & \multirow{2}{*}{\claude{} Sonnet-4.6} & \noiseoff & \textbf{\heat{72.6}{38.1}{72.6}\%} & \heat{75.6}{40.4}{78.5}\% & \textbf{\heat{60.7}{26.7}{60.7}\%} & 292.5 & 702.0 & 1.47M \\
                                     &                                    & \noiseon & \heat{62.6}{38.1}{72.6}\% & \heat{76.3}{40.4}{78.5}\% & \heat{53.7}{26.7}{60.7}\% & 314.0 & 736.5 & 1.71M \\
        \midrule
        \multirow{2}{*}{Codex} & \multirow{2}{*}{\openai{} GPT-5.4} & \noiseoff & \heat{70.0}{38.1}{72.6}\% & \heat{63.7}{40.4}{78.5}\% & \heat{53.3}{26.7}{60.7}\% & 176.4 & \textbf{376.0} & 1.98M \\
                               &                                 & \noiseon  & \heat{59.3}{38.1}{72.6}\% & \heat{61.9}{40.4}{78.5}\% & \heat{45.9}{26.7}{60.7}\% & 218.1 & 397.7 & 1.88M \\
        \bottomrule
    \end{tabular}
\end{table*}

Table~\ref{tab:main-results} shows the overall results.
\bench{} presents challenges to frontier AI agents
    and models, 
    with diagnosis success rates ranging from 38.1\% to 72.6\%
    and mitigation success rates ranging from 40.4\% to 78.5\%
    across agent-model pairs.
The SRE problems evaluate
    an agent's ability to reason about complex interactions between system components,
    infer root causes from symptoms,
    and effectively use tools to operate systems,
    all of which remain difficult for frontier agents and models.

We find that agents show different characteristics in addressing SRE problems.
Claude Code achieves the highest end-to-end success rates, compared to 
    Stratus and Codex.
Stratus with Sonnet-4.6 has 
    the highest mitigation success rate among all agents, attributed
    to its undo-and-retry mechanism~\cite{chen:nips:25}.
On the other hand, Stratus with Kimi K2.5 has the lowest success rates,
    due to the weaker capabilities of the underlying model.

{\bf Token Cost.}
Claude Code uses 1.81$\times$ more tokens per run than Stratus with the same model, 
    and Codex uses at least 2.44$\times$ more.
The reason is that coding agents are not optimized for processing
    the large volumes of observability data in SRE tasks.
Stratus, as an SRE agent, preprocesses observability data and only prompts
    LLMs with relevant data.


{\bf Impact of Noises.}
As shown in Table~\ref{tab:main-results},
    diagnosis success rate drops for {\it every} agent-model pair.
Mitigation success is more robust than diagnosis under noises.
Noises mislead an agent's hypotheses
    about the root causes; 
    on the other hand, agents tend to recover from incorrect hypotheses with
    self-validation (see \S\ref{sec:eval-diag-mit-correlation}).
Inspecting trajectories,
    we observe that all the evaluated agents take a {\it greedy} approach 
    (see Appendix~\ref{sec:case}):
    they always treat the first plausible anomaly as
    the target failure (which can be a noise).
In several cases, the agents did find evidence of the root
    cause of the target failure but disregarded it because
    it was unrelated to the noise they were (wrongly) targeting.



\subsection{Results on New Failure Scenarios}
\label{sec:eval-novel-faults-noise}



Table~\ref{tab:shared-vs-novel} partitions \bench{}'s problems by failure scenarios.
We find that problems ported from existing benchmarks~\cite{chen:mlsys:25,jha:icml:25} pose 
    limited challenges for the evaluated agents with strong models.
For example, the mitigation success rate is above 80\% for Stratus with Sonnet-4.6.
Most of these problems are single-fault scenarios focusing on applications and do not include noises. 
\bench{} includes {\it new} problems that are built on 
    similar failure scenarios in terms of fault families,
    but differ in applications and system components.
As shown in Table~\ref{tab:shared-vs-novel}, evaluated agents show 
    similar results.

However, Table~\ref{tab:shared-vs-novel} shows that the agents
    perform significantly worse on new failure scenarios that 
    are unique to \bench{}.
The end-to-end success rates of Stratus with Sonnet-4.6, Claude Code, and Codex
    decrease from 63.7\% to 17.9\%, 60.8\% to 28.2\%, and 57.8\% to 15.4\%, respectively.
These results show significant gaps in AI agents' ability to address 
    high-fidelity failures such as those rooted in low-level stacks
    and/or caused by compound failures.
We briefly describe two case studies, with more details in Appendix~\ref{sec:case}. 

{\bf Hardware faults.}
\bench{}'s {\tt latent\_sector\_error} problem injects intermittent
    errors into \code{read} system calls on a node.
Across three runs of Stratus and Claude Code, no run
    produced an aggregated diagnosis score above $0.22$, and
    fault characterization received a score of $0$ in every run.
All agents attribute the errors to error-handling logic
    inside the application.
None of the runs proposed disk-level diagnostics
    (e.g., reading {\tt dmesg} and inspecting {\tt smartctl} output).

{\bf Metastable failures.}
The evaluated agents reliably diagnose the application-level trigger, which
    is visible in distributed traces and deployment configurations;
however, no agent across the metastable failure problems
    identified {\it both} interacting components.
In one run, Codex did locate the resource constraint but
    then dismissed the application trigger as a downstream
    artifact, producing a diagnosis that was correct on
    localization but wrong on scope (\S\ref{sec:oracle}).

In general, we observe a lack of coherent,
    comprehensive understanding of the interactions among the control/management
    plane of the cluster, the deployed systems, and the
    user requests.
On metastable failures, agents did not connect the 
    system-level constraints with the application-level
    trigger: this is required to identify how the system
    is induced into the metastable state.
On hardware failures, agents also miss the
    hardware-software interactions.
This lack of understanding limits agents' ability to
    diagnose complex failures in \bench{}, which are arguably closer to real-world incidents.

\providecommand{\cmark}{\ding{51}}
\providecommand{\xmark}{\ding{55}}
\begin{table}[t]
    \caption{Benchmark results partitioned into three
        problem types. ``Ported'' refers to problems directly
        ported from AIOpsLab/ITBench; ``Similar Failures'' are new 
        problems in \bench{} that share failure patterns with ``Ported'';
        ``New failures'' are faults and failure modes unique to \bench{}.}
    \label{tab:shared-vs-novel}
    \vspace{3pt}
    \centering
    \footnotesize
    \setlength{\tabcolsep}{6pt}
    \begin{tabular}{lc|ccc|ccc|ccc}
        \toprule
        & & \multicolumn{3}{c|}{\textbf{Ported} ($n{=}34$)}
          & \multicolumn{3}{c|}{\textbf{Similar Failures} ($n{=}43$)}
          & \multicolumn{3}{c}{\textbf{New Failures} ($n{=}13$)} \\
        \cmidrule(lr){3-5}\cmidrule(lr){6-8}\cmidrule(lr){9-11}
        Agent & Noise & Diag. & Mitig. & E2E
                      & Diag. & Mitig. & E2E
                      & Diag. & Mitig. & E2E \\
        \midrule
        \multirow{2}{*}{Stratus \claude{}}
            & \noiseoff & \heat{70.6}{39.2}{76.5}\% & \textbf{\heat{83.3}{46.1}{83.3}\%} & \textbf{\heat{63.7}{27.5}{63.7}\%} & \heat{62.8}{43.4}{81.4}\% & \textbf{\heat{80.6}{41.9}{80.6}\%} & \heat{58.9}{30.2}{70.5}\% & \heat{33.3}{15.4}{51.3}\% & \heat{59.0}{12.8}{76.9}\% & \heat{17.9}{10.3}{48.7}\% \\
            & \noiseon & \heat{58.8}{39.2}{76.5}\% & \heat{68.6}{46.1}{83.3}\% & \heat{45.1}{27.5}{63.7}\% & \heat{55.0}{43.4}{81.4}\% & \heat{64.3}{41.9}{80.6}\% & \heat{44.2}{30.2}{70.5}\% & \heat{20.5}{15.4}{51.3}\% & \heat{30.8}{12.8}{76.9}\% & \heat{10.3}{10.3}{48.7}\% \\
        \midrule
        \multirow{2}{*}{Stratus \kimi{}}
            & \noiseoff & \heat{42.2}{39.2}{76.5}\% & \heat{46.1}{46.1}{83.3}\% & \heat{27.5}{27.5}{63.7}\% & \heat{46.5}{43.4}{81.4}\% & \heat{44.2}{41.9}{80.6}\% & \heat{32.6}{30.2}{70.5}\% & \heat{15.4}{15.4}{51.3}\% & \heat{12.8}{12.8}{76.9}\% & \heat{10.3}{10.3}{48.7}\% \\
            & \noiseon & \heat{39.2}{39.2}{76.5}\% & \heat{49.0}{46.1}{83.3}\% & \heat{27.5}{27.5}{63.7}\% & \heat{43.4}{43.4}{81.4}\% & \heat{41.9}{41.9}{80.6}\% & \heat{30.2}{30.2}{70.5}\% & \heat{17.9}{15.4}{51.3}\% & \heat{23.1}{12.8}{76.9}\% & \heat{12.8}{10.3}{48.7}\% \\
        \midrule
        \multirow{2}{*}{Claude Code}
            & \noiseoff & \heat{74.5}{39.2}{76.5}\% & \heat{71.6}{46.1}{83.3}\% & \heat{60.8}{27.5}{63.7}\% & \textbf{\heat{81.4}{43.4}{81.4}\%} & \heat{79.1}{41.9}{80.6}\% & \textbf{\heat{70.5}{30.2}{70.5}\%} & \heat{38.5}{15.4}{51.3}\% & \heat{74.4}{12.8}{76.9}\% & \heat{28.2}{10.3}{48.7}\% \\
            & \noiseon & \heat{66.7}{39.2}{76.5}\% & \heat{73.5}{46.1}{83.3}\% & \heat{52.0}{27.5}{63.7}\% & \heat{62.8}{43.4}{81.4}\% & \heat{78.3}{41.9}{80.6}\% & \heat{56.6}{30.2}{70.5}\% & \textbf{\heat{51.3}{15.4}{51.3}\%} & \textbf{\heat{76.9}{12.8}{76.9}\%} & \textbf{\heat{48.7}{10.3}{48.7}\%} \\
        \midrule
        \multirow{2}{*}{Codex}
            & \noiseoff & \textbf{\heat{76.5}{39.2}{76.5}\%} & \heat{66.7}{46.1}{83.3}\% & \heat{57.8}{27.5}{63.7}\% & \heat{78.3}{43.4}{81.4}\% & \heat{68.2}{41.9}{80.6}\% & \heat{61.2}{30.2}{70.5}\% & \heat{25.6}{15.4}{51.3}\% & \heat{41.0}{12.8}{76.9}\% & \heat{15.4}{10.3}{48.7}\% \\
            & \noiseon & \heat{66.7}{39.2}{76.5}\% & \heat{66.7}{46.1}{83.3}\% & \heat{51.0}{27.5}{63.7}\% & \heat{62.8}{43.4}{81.4}\% & \heat{68.2}{41.9}{80.6}\% & \heat{50.4}{30.2}{70.5}\% & \heat{28.2}{15.4}{51.3}\% & \heat{28.2}{12.8}{76.9}\% & \heat{17.9}{10.3}{48.7}\% \\
        \bottomrule
    \end{tabular}%
\end{table}

\subsection{Correlations between Diagnosis and Mitigation}
\label{sec:eval-diag-mit-correlation}

\begin{figure}[t]
\centering
\begin{minipage}[t]{0.40\columnwidth}
    \centering
    \vspace{0pt}
    \captionof{table}{Conditional probability of mitigation success (M) given the diagnosis outcome (D).}
    \label{fig:diag-mit-analysis}
    \footnotesize
    \footnotesize
\setlength{\tabcolsep}{3pt}
\begin{tabular}{llcc}
    \toprule
    Agent & Noise & $P(\text{M} \mid \text{D})$ & $P(\text{M} \mid \neg \text{D})$ \\
    \midrule
    Stratus \claude{}  & \noiseoff & 0.892 & 0.615 \\ 
    Stratus \claude{}  & \noiseon & 0.770 & 0.443 \\ 
    Stratus \kimi{}    & \noiseoff & 0.679 & 0.217 \\ 
    Stratus \kimi{}    & \noiseon & 0.699 & 0.246 \\ 
    Claude Code        & \noiseoff & 0.836 & 0.541 \\ 
    Claude Code        & \noiseon & 0.858 & 0.604 \\ 
    Codex              & \noiseoff & 0.762 & 0.346 \\ 
    Codex              & \noiseon & 0.775 & 0.391 \\ 
    \bottomrule
\end{tabular}

%
\end{minipage}%
\hfill
\begin{minipage}[t]{0.58\columnwidth}
    \centering
    \vspace{0pt}
    \includegraphics[width=\linewidth]{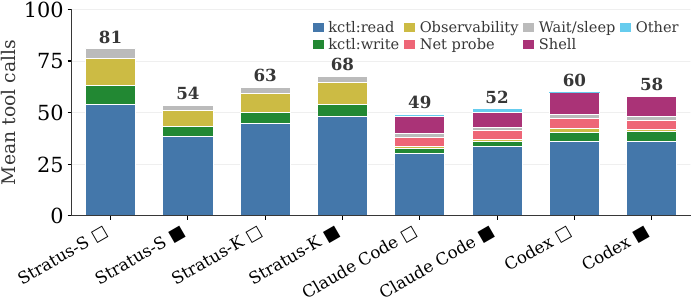}
    \captionof{figure}{The average number of 
        tool calls per SRE problem by category. 
        S: Sonnet~4.6, K: Kimi~K2.5; $\blacksquare$: noises.}
    \label{fig:tool-usage}
\end{minipage}
\end{figure}

When an agent succeeds at mitigation, was that success
     backed by a correct diagnosis, or did the agent stumble into a
     fix without knowing why?
Table~\ref{fig:diag-mit-analysis} shows 
    conditional probabilities of mitigation given diagnosis outcome.
Agents may pinpoint the root cause but fail in mitigation:
in cases where the agent correctly diagnoses the failure,
    mitigation succeeds only 68\%--89\% of the time
    ($P(\text{M} \mid \text{D})$).
When the initial diagnosis fails to address the root
    cause, mitigation success rate drops by 25--46 percentage points
    across agents.
The drop shows that diagnosis is helpful in
    guiding the agent in its mitigation.



When the diagnosis is 
    wrong, the agent may still 
    successfully mitigate the failure by continuously
    observing and tuning the systems.
Mitigation succeeds 22--62\% of the time across agents when
    the initial diagnosis is incorrect ($P(\text{M} \mid \neg \text{D})$).
Note that the benchmark does not inform the agent whether its
    diagnosis is correct or not.
We observe two common patterns:
    (1) the agent often pattern-matches a known symptom
        to mitigation actions that can successfully
        mitigate the failure without understanding its root cause;
    (2) the agent can correct its
        incorrect diagnosis by observing the persistence of the failure 
        and continuously forming new hypotheses to
        make further attempts.
For example, without noises injected, when the diagnosis is incorrect, Stratus with
    Sonnet-4.6 averages 3.82 attempts during mitigation, compared with
    1.88 attempts with correct diagnosis
    (see Table~\ref{tab:tool-retries}).

\vspace{-2pt}
\subsection{Tool Usage}
\label{sec:tool-usage}

Figure~\ref{fig:tool-usage} classifies tool calls in the agent trajectories
    (Appendix~\ref{sec:appendix:tool-usage} shows more details).
The majority (60--72\%)
     of tool calls are read-only \code{kubectl} commands.
Two commands account for around 87\% of all read operations:
    \code{kubectl} \code{get} to inspect system state
    and 
    \code{kubectl} \code{logs} to read container logs.
The primary resources inspected are Pods~\cite{pod}, Deployments~\cite{deployment},
    and ConfigMaps~\cite{configmap}.
Agents execute about 19--28 read commands before their first mitigation action
    (see Table~\ref{tab:tool-rw-ratio}).

Agents differ in their mitigation actions.
Stratus favors \code{kubectl} \code{patch} (39--41\% of write actions)
    to change system state,
    while Claude Code and Codex prefer
    \code{kubectl} \code{rollout} (29--40\% of write actions),
    which restarts deployments.
Codex additionally relies on \code{kubectl} \code{run} (19--21\%)
    to spin up ephemeral containers for in-cluster network testing
    and \code{kubectl} \code{port-forward}
    (11--12\%) for direct service probing,
    which Stratus and Claude Code rarely use.

Stratus has custom-built API tools for querying
    metrics, traces, and service
    dependency graphs, which account for 15--17\%
    of its tool calls.
Claude Code and Codex instead 
    discover observability endpoints and construct
    raw \code{curl} commands;
    their observability access accounts for only
    2--3\% of tool calls.

\section{Engineering Practices for SRE Benchmarks}
\label{sec:engineering-practices}

We recommend the following practices for future SRE benchmarks;
    each addresses a pitfall in existing ones~\cite{chen:mlsys:25,jha:icml:25,zhang:arxiv:25,cloudopsbench}.

{\bf Evaluate in live environments, not snapshots.}
Presenting telemetry as a static snapshot~\cite{cloudopsbench} removes
    nondeterminism, but partial signals, noise, and evolving state are
    intrinsic to production.
A snapshot reduces troubleshooting to log triage and removes the skills
    the benchmark should measure: forming hypotheses, issuing probes, and
    observing how the system responds over time.

{\bf Do not constrain the agent architecture.}
AIOpsLab~\cite{chen:mlsys:25} mediates every action through a
    ReAct~\cite{react} orchestrator, so agents with other designs must be ported to it.
\bench{} requires only a \texttt{submit()} call, so it evaluates Stratus, Claude Code, and Codex alike.

{\bf Make the runtime programmable.}
A metastable failure (\S\ref{sec:problems}) requires injecting a trigger and
    a resource constraint with the right timing and ordering, plus an oracle that keeps
    observing the system to distinguish recovery from relapse;
    noise requires a concurrent injection loop.
Substrates such as the Ansible Playbooks in ITBench~\cite{jha:icml:25}
    push this coordination into {\it ad hoc} shell scripts that are hard to control and reproduce.
\bench{} schedules faults, noises, and oracle probes in a single Python service,
    so timing-sensitive behavior is expressed in the problem definition itself.

{\bf Inject defects, not symptoms.}
Chaos-engineering tools~\cite{chaosmesh,chaosblade} inject symptoms
    (killed pods, dropped packets, throttled CPUs) rather than defects,
    so the only ``mitigation'' is to stop the injector.
Yet MicroRemed~\cite{zhang:arxiv:25} sources all of its faults from Chaos Mesh,
    Cloud-OpsBench~\cite{cloudopsbench} draws on Chaosblade,
    and 6 of the 36 ITBench SRE scenarios wrap a Chaos Mesh schedule.
\bench{} injects faults via direct Kubernetes manipulation, eBPF-based syscall
    failures, and operator misoperations, so every problem has a real defect
    to diagnose and resolve.

{\bf Guard against reward hacking.}
Agents can exploit benchmark infrastructure instead of solving the task~\cite{benchjack}.
In AIOpsLab and ITBench, fault injectors run as identifiable pods that the
    agent can find and disable;
    moreover, 8 of 18 ITBench mitigation problems can be ``solved'' by a
    generic pod-restart loop that clears the alert without fixing the defect~\cite{chen:nips:25}.
\bench{} hides its fault-injection plane behind a proxy that agents cannot see,
    and its mitigation oracles probe live system state, not alerts.

\section{Related Work}
\label{sec:related-work}

The advances of AI for SWE (Software Engineering) have pushed 
    AI for SRE to be the next frontier.
Recent work has made active progress on agentic SRE technologies,
    from root cause analysis (RCA) to failure mitigation~\cite{chen:eurosys:24,pei:www:25,wang:cikm:24,zhang:emnlp:2024,zhang:tsc:25,tang:arxiv:25,luo:arxiv:25,zhang:arxiv:25:adaptive,mao:arxiv:25,zhong:tsc:26,chen:nips:25}; meanwhile,
    many commercial and open-source SRE agent products have been developed~\cite{ciroos,devin-oncall,azure-sre-agent,aws-sre-agent}.
Our communications with many SRE researchers and practitioners show that 
    high-quality SRE benchmarks are highly desired.

A few benchmarks provide static datasets of Q\&A~\cite{liu:arxiv:25,sre-skills-bench},
    observability data from real/synthetic sources~\cite{han:nips:22,jacob:vldb:21,xu:iclr:25},
    and system snapshots~\cite{cloudopsbench}.
\bench{} can also be used to generate such datasets.
These datasets are valuable for anomaly detection and RCA,
    but are fundamentally limited as they do not provide opportunities for agents 
    to iteratively probe and observe the systems as in real-world diagnosis.
For the same reason, these benchmarks cannot support mitigation tasks.

The design of \bench{} stems from reflection on 
    existing live benchmarks~\cite{chen:mlsys:25,jha:icml:25,zhang:arxiv:25}
    which are limited in their abilities to simulate
    high-fidelity, realistic failure scenarios, 
    due to the lack of system support for orchestrating
    distributed events, low-level fault simulation and injection mechanisms, 
    noise simulation, etc.
They also suffer from the engineering pitfalls discussed in \S\ref{sec:engineering-practices}.

\bench{} focuses on failure analysis and mitigation and thus is complementary to 
    other related benchmarks for deployment and regression testing~\cite{cloudopsbench} 
    and terminal-based environment setup~\cite{merrill:arxiv:26}.

\section{Conclusion}
\label{sec:discussion}

The aspiration of \bench{} is to push the standard
    of AI for SRE
    and to unlock a whole new evolution of agentic SRE technologies.
With this goal in mind, we develop \bench{} as a high-fidelity 
    benchmark to represent existing challenges in real-world production 
    environments
    and as a usable, extensible
    framework that can be continuously evolved with new 
    challenges for AI.
Several components of \bench{} can be further enhanced and enriched,
    including more diverse noise modeling, more comprehensive fault simulation
    and failure modes, and system environments (e.g., to include edge presences)---some of which are already underway.
A clear next step is to upgrade \bench{} into a reinforcement-learning style 
    training ground for SRE agents.

\begin{ack}

We thank everyone who has supported, helped, and contributed to \bench{}.
Specifically, we thank all the contributors to \bench{}
    and all the users of \bench{} who have given us encouragement and feedback.
We thank Braden Hancock, Andy Konwinski, Brighten Godfrey, Sasa Misailovic, Xuan Feng,
    Lidong Zhou, and Darko Marinov
    for valuable feedback on the project.

\bench{} is supported in part by a Slingshot grant from the Laude Institute,
    the ChatGPT for Academic Researchers program, 
    the Anthropic AI for Science program,
    E2B for Research,
    Cohere Labs Catalyst Grant 
    and NSF CNS-2145295.

{\bf Competing interests.} The authors declare no competing interests.

\end{ack}

\bibliographystyle{abbrvnat}
\bibliography{ref}

\newpage
\appendix
\section{Diagnosis Evaluation Checklist}
\label{sec:appendix:checklist}

Table~\ref{tab:checklist} lists the full checklist used by the
    diagnosis oracle (\S\ref{sec:oracle}).
Each question requires a Yes/No answer;
    the LLM evaluator returns supporting evidence
    and a confidence (High/Medium/Low) per question.
Dimension weights and the pass threshold are set in
    a YAML file (current defaults: $w = \frac{1}{3}$ per dimension and $\tau = \frac{7}{9}$). 

\begin{table}[h]
\centering
\small
\setlength{\tabcolsep}{4pt}
\caption{Full diagnosis evaluation checklist.
    Each question is answered Yes/No by the LLM evaluator.
    The per-dimension score formula and the aggregated
    score formula can be found in \S\ref{sec:oracle}.}
\vspace{5pt}
\begin{tabular}{p{0.16\textwidth} p{0.08\textwidth} p{0.42\textwidth} p{0.24\textwidth}}
\toprule
\textbf{Dimension} & \textbf{ID} & \textbf{Question} & \textbf{Evaluator Hint} \\
\midrule
\multirow{3}{*}{\parbox{0.16\textwidth}{Fault\\Localization\\($w\!=\!\frac{1}{3}$)}} & D1-Q1
  & Does the diagnosis name the same service, deployment, pod, node, or infrastructure component that the ground-truth identifies as the fault origin?
  & Compare against the target component and target resource type in the fault specification (spec). \\
\cmidrule(lr){2-4}
 & D1-Q2
  & Does the diagnosis correctly distinguish the fault origin from any secondary or cascading failure points mentioned in the ground-truth?
  & Check that the diagnosis points to the root-cause component, not a downstream victim. \\
\cmidrule(lr){2-4}
 & D1-Q3
  & Does the diagnosis avoid misidentifying a healthy component as the fault origin?
  & Verify the diagnosed component matches the fault spec's target component. \\
\midrule
\multirow{3}{*}{\parbox{0.16\textwidth}{Fault\\Characterization\\($w\!=\!\frac{1}{3}$)}} & D2-Q1
  & Does the diagnosis identify the same injected mechanism described in the ground-truth (e.g., wrong network port, missing environment variables, wrong container image, wrong selector, or wrong memory limit)?
  & Match against the fault mechanism and injector method in the structured spec. \\
\cmidrule(lr){2-4}
 & D2-Q2
  & Does the diagnosis include concrete mutated details from the injection logic (e.g., environment variable, configuration value, network port, selector, container image tag, or resource limit)?
  & Compare concrete claims against parameters and the target mutation implied by the injector method. \\
\cmidrule(lr){2-4}
 & D2-Q3
  & Does the diagnosis avoid attributing the fault to an incorrect or unrelated fault type?
  & Check that the diagnosis does not conflict with the problem class, injector method, or injected parameter values. \\
\midrule
\multirow{3}{*}{\parbox{0.16\textwidth}{Scope\\Precision\\($w\!=\!\frac{1}{3}$)}} & D3-Q1
  & Does the diagnosis avoid blaming components that are not identified in the ground-truth as contributing to the fault?
  & Check for over-attribution: the diagnosis should not blame uninvolved components. \\
\cmidrule(lr){2-4}
 & D3-Q2
  & Does the diagnosis include all components listed in the ground-truth as contributing to or affected by the fault?
  & Check for under-attribution: all ground-truth components should be pointed out. \\
\cmidrule(lr){2-4}
 & D3-Q3
  & Does the diagnosis correctly describe the impact or symptom consistent with what the ground-truth states?
  & Compare stated impact against mechanism, parameters, and target component in the fault spec. \\
\bottomrule
\end{tabular}
\label{tab:checklist}
\end{table}

\section{Combinatorial Coverage Details}
\label{sec:appendix:coverage}

We provide the detailed breakdown behind the
    combinatorial coverage reported in the preamble of \S\ref{sec:problems}.
Table~\ref{tab:appendix:fault-target-compat} groups \bench{}'s
    fault primitives by the classes of target components (referred to as
    ``target'' for short) they are
    compatible with, and reports the number of viable
    (fault, target) pairs per class across the 139-pod injection-target space
    (plus three worker nodes for hardware-level faults).

\begin{table}[H]
\caption{Fault-target compatibility classes. ``\# Faults'' refers to the
    number of fault primitives in each class; since one fault primitive
    can be compatible with multiple target classes (e.g., a missing
    environment variable affects both MongoDB and non-MongoDB pods), the
    column does not sum to the primitive total. ``Pairs'' is the count of
    viable (fault, target) combinations, totaling 3{,}623.}
\label{tab:appendix:fault-target-compat}
\vspace{3pt}
\centering
\small
\setlength{\tabcolsep}{4pt}
\begin{tabular}{lrlr}
\toprule
\textbf{Class} & \textbf{\# Faults} & \textbf{Compatible targets} & \textbf{Pairs} \\
\midrule
Universal Kubernetes-level      & 25 & 139 pods                          & 3{,}475 \\
Storage-dependent               & 5  & 6 PVC-mounted pods                & 30 \\
DaemonSet-level                 & 1  & 3 DaemonSets                      & 3 \\
Operator-level                  & 6  & 5 Kubernetes applications         & 30 \\
MongoDB-specific                & 4  & 18 MongoDB pods                   & 72 \\
Valkey-specific                 & 2  & 1 pod                             & 2 \\
App-layer misconfiguration      & 1  & 2 services                        & 2 \\
Node/kernel                     & 3  & 3 worker nodes                    & 9 \\
\midrule
\textbf{Total}                  &    &                                   & \textbf{3{,}623} \\
\bottomrule
\end{tabular}
\vspace{3pt}
\end{table}

The 90 curated problems we evaluate against in \S\ref{sec:eval-setup}
    exercise only 2.5\% of the 3{,}623 viable
    (fault, target) pairs.
The noise simulation and multi-fault composition
    further multiply the combinations.
The same combinatorial structure positions \bench{} as a natural
    environment for reinforcement-learning rollouts, where each
    (fault, target) pair becomes an episode against a live, production-like system environment
    (\S\ref{sec:discussion}).

\section{Case Studies}
\label{sec:case}

\subsection{Metastable Failure}
\label{sec:case-metastable}

\bench{} includes problems that model
    metastable failures,
    where a temporary trigger pushes the system into
    a self-sustaining degraded state that persists
    even after the trigger is removed~\cite{bronson:hotos:21,huang:osdi:22,Isaacs:hotos:25}.
Each problem is a {\it compound fault}:
    an application-level trigger
    (e.g., a misconfigured retry policy that amplifies
    traffic, or a runtime flag that forces
    frequent garbage collection)
    paired with an infrastructure constraint that drives the system
    into a vulnerable state
    (e.g., a resource quota or limit that caps CPU or memory
    allocation).
The constraint produces no errors and no failed traces;
    it is discoverable only through explicit inspection
    of deployment configuration.
The trigger drives the system from the vulnerable state
    to the self-sustaining metastable state: 
    system performance degrades
    with no explicit fail-stop symptoms.
Fixing only the trigger does not mitigate
    the symptoms: the agent must reason about 
    the relationship between
    the trigger and the sustained degraded state,
    then mitigate the problem by removing the 
    infrastructure-level constraint and
    restarting the application,
    giving it a clean slate.

In our evaluation, agents reliably diagnosed the
    application-level trigger through trace analysis
    and deployment inspection,
    but almost never discovered the infrastructure
    constraint because it produces no observable symptom.
In the one run where Codex {\it did} find the
    infrastructure resource constraint, it attributed the entire
    failure to it and dismissed the application-level
    misconfiguration.
In no case did any agent identify the interaction
    between the trigger and the constraint.

Agents do occasionally mitigate the metastable behavior.
Agents fixed the misconfiguration and restarted
    the affected services;
    this restored the system and cleared the amplified traffic,
    so the system can resume normal execution.
When the agent fails to mitigate the metastable behavior,
    it is often because the agent is distracted by
    surface-level symptoms and attributes the failure
    to an entirely different fault type.
For example, in one run of the {\tt gc\_capacity\_degradation} problem,
    where an aggressively low {\tt GOGC} setting forces
    frequent garbage collection across all workloads in a
    capacity-constrained namespace,
    the agent never inspected {\tt GOGC} at all.
Instead, the agent fixated on distributed traces showing
    a $\sim$300\,ms gap between the frontend and the
    profile-service, while the server-side
    {\tt GetProfiles} handler completed in microseconds.
From this pattern alone, the agent concluded that the
    delay must live in the network path, and submitted a
    root cause of a {\tt tc} {\tt netem} rule injecting
    $\sim$150\,ms of latency in each direction on the
    profile-service pod's {\tt veth} interface.
This incorrect diagnosis makes two independent errors:
    (1) it misidentifies the fault type as network latency injection, and
    (2) it limits the scope to a single service.
As a result, the agent proposed mitigations targeting a
    nonexistent network rule, and never restarted
    the workloads to clear the metastable 
    garbage-collection loop.

\subsection{Hardware Fault}

The {\tt latent\_sector\_error} problem in \bench{} injects
    hardware faults into the storage devices of a physical node.
The faulty disk returns intermittent errors
    on file-reading system calls (e.g., \texttt{pread()}).
The MongoDB databases deployed on that node crash with
    ``\code{read:} \code{input/output} \code{error},'' as they are unable to
    read storage-engine-related metadata files.

Our evaluation shows that agents struggle to diagnose the hardware-related
    root cause.
Across three Stratus, three Claude Code, and three
    Codex runs with
    no noise simulation,
    none produced a diagnosis score above $0.22$,
    and D2 (fault characterization) scored $0$ in every run,
    showing that their diagnoses never identified the correct
    fault mechanism.
For example, the agents observe that the I/O errors come from the \texttt{memcached}
    deployment, which reads from the MongoDB deployment, and incorrectly
    attribute the failure to dropped connections from the \texttt{memcached} deployment.
Agents also blame user workloads for overwhelming the \texttt{memcached}
    process, causing it to reset connections.
Lastly, they blame the microservice application for not gracefully
    handling the I/O error.
These conclusions never mention that the underlying
    hardware can be defective.

This problem exposes a specific weakness:
    agents lack an understanding of how 
    the underlying hardware
    can affect the application deployed atop it.
Agents treat the exposed I/O error as evidence of 
    incorrect error handling, but 
    never suggest disk-level diagnostics that 
    are standard SRE responses
    to hardware failures.

\subsection{Greedy Approach in Diagnosing Failures}

A recurring agent failure pattern across our evaluation
    is the agent taking a greedy approach in 
    diagnosing and mitigating the first observed anomaly in the environment.
The agent first encounters a suspected but unrelated 
    anomaly in the environment,
    and submits a premature diagnosis,
    which later misleads the mitigation phase.
These suspected anomalies can be injected noise
    running alongside the target fault,
    or pre-existing application characteristics
    such as low memory limits in application containers.

We observe that the pattern is consistent regardless 
    of the anomaly the agent picks up.
Figure~\ref{fig:intro-trace} shows an example using
    the {\tt missing\_env\_variable\_astronomy\_shop} problem,
    where the actual fault is a dropped environment variable
    on the frontend component of the application.
Across all runs of both Stratus and Claude Code,
    neither agent ever inspected the frontend component.
Stratus latched onto another deployment
    with a low memory limit,
    while Claude Code focused on
    the Jaeger tracing infrastructure.
Both found plausible-looking issues,
    diagnosed them as the root cause,
    and never investigated the actual faulty component.
This greedy approach affects mitigation effectiveness.
After fixating on the memory limit, the agent raises it
    with \texttt{kubectl} \texttt{patch} or a similar edit,
    but clearly this fix does not address the offending fault.

A more interesting (and arguably more troubling) case occurs when
    the agent {\it does} encounter evidence of the
    offending fault but still anchors on its
    initial hypothesis.
In the {\tt valkey\_auth\_disruption} problem, 
    the fault is an invalid password set on the
    Valkey database at runtime.
The dependent microservice component crashes 
    because it cannot
    authenticate to the database.
Agents often attributed the crash to insufficient memory,
    noting that the containers had
    low memory limits.
However, the agents also observed that
    the init container's TCP health check succeeded
    while the application-layer database connection failed.
This is a discrepancy consistent with authentication failure,
    not resource exhaustion.
In one run, the agent mentioned the password-setting command ({\tt requirepass})
    multiple times as a candidate hypothesis
    but defaulted to the memory explanation.

This case study exposes a specific weakness:
    the agent treats the first plausible abnormality as a
    stopping criterion.
Once a candidate root cause is written down,
    in the mitigation phase, the model
    interprets subsequent evidence as supporting it and
    does not generate competing hypotheses.
A human SRE, in contrast, would be able to form multiple
    hypotheses and investigate concurrently, until the real
    root cause is found.


\section{Tool Usage Details}
\label{sec:appendix:tool-usage}

We provide a detailed analysis of the
    tool usage reported in \S\ref{sec:tool-usage}.
We classify every tool call in the agent trajectories into
    the following categories:
\begin{itemize}[leftmargin=*,topsep=-2pt,itemsep=1pt]
    \item \textbf{kubectl (read)} for read-only inspection of system state
        (e.g., \code{get}, \code{logs}, \code{describe},
        \code{top}, \code{auth}).
    \item \textbf{kubectl (write)} for changing system state
        (e.g., \code{patch}, \code{rollout}, \code{set}, \code{run},
        \code{delete}, \code{apply}, \code{scale},
        \code{port-forward}).
    \item \textbf{Observability} for querying metrics, traces, and
        service maps.
        Stratus uses dedicated agent tools;
        Claude Code and Codex use \code{curl} to query
        Prometheus, Jaeger, or Loki endpoints.
    \item \textbf{Network} for connectivity probes
        (e.g., \code{curl} to application endpoints,
        \code{ss}, \code{netstat}, \code{nc}, \code{ping}).
    \item \textbf{Wait/sleep} for deliberate pausing
        (e.g., \code{wait\_tool} for Stratus and \code{sleep} for others).
    \item \textbf{Shell} for general-purpose utilities
        (e.g., \code{cat}, \code{ls}, \code{grep}, \code{python},
        \code{sed}, \code{export}).
    \item \textbf{Others} such as agent-framework-native tools
        (e.g., Claude Code's \code{Read}, \code{Agent},
        \code{WebSearch}; Kimi's \code{thinking} tool).
\end{itemize}
\noindent
Submission calls are excluded
    from all counts.
Table~\ref{tab:tool-categories} shows the tool 
    call category breakdown per
    evaluation configuration.
Table~\ref{tab:tool-rw-ratio} shows the ratio of
    read to write commands in the system environment.
Table~\ref{tab:tool-top-reads} and 
    Table~\ref{tab:tool-top-writes} show the top
    subcommands executed in \texttt{kubectl}.
All averages are computed as the mean across runs per problem,
    then the mean across problems.

\begin{table}[ht]
\centering
\footnotesize
\setlength{\tabcolsep}{3.5pt}
\caption{Tool call category breakdown.}
\label{tab:tool-categories}
\vspace{3pt}
\begin{tabular}{llcccccccc}
\toprule
Agent & Model & Noise & kctl(read) & kctl(write) & Observability & Network & Wait & Shell & Other \\
\midrule
\multirow{4}{*}{Stratus} & \multirow{2}{*}{\claude{} Sonnet 4.6} & \noiseoff & 65.8 & 11.9 & 16.6 & 0.0 & 5.6 & 0.0 & 0.0 \\
                         &                                    & \noiseon & 71.8 &  8.3 & 14.5 & 0.0 & 5.1 & 0.0 & 0.2 \\
                         \cmidrule{2-10}
                         & \multirow{2}{*}{\kimi{} K2.5}      & \noiseoff & 71.6 &  8.5 & 14.6 & 0.0 & 4.6 & 0.0 & 0.7 \\
                         &                                    & \noiseon & 72.0 &  8.1 & 14.8 & 0.0 & 4.4 & 0.0 & 0.6 \\
\midrule
\multirow{2}{*}{Claude Code} & \multirow{2}{*}{\claude{} Sonnet 4.6} & \noiseoff & 61.5 &  4.6 &  2.4 & 8.3 & 4.2 & 16.5 & 2.5 \\
                             &                                    & \noiseon & 64.8 &  4.5 &  2.1 & 8.5 & 2.8 & 14.0 & 3.4 \\
\midrule
\multirow{2}{*}{Codex} & \multirow{2}{*}{\openai{} GPT-5.4} & \noiseoff & 60.2 &  7.7 &  2.4 & 8.7 & 2.7 & 18.4 & 0.1 \\
                       &                                  & \noiseon & 61.6 &  7.6 &  2.7 & 7.4 & 2.7 & 17.8 & 0.2 \\
\bottomrule
\end{tabular}
\end{table}

\begin{table}[ht]
\centering
\small
\caption{\code{kubectl} read/write analysis.
    Ratio is total reads divided by total writes.
    ``Reads $\to$ First Write'' is the mean number of read-only
    \code{kubectl} commands before the first mutation
    of system state,
    averaged per problem then across problems.}
\label{tab:tool-rw-ratio}
\vspace{3pt}
\begin{tabular}{llccc}
\toprule
Agent & Model & Noise & Ratio & Reads $\to$ First Write \\
\midrule
\multirow{4}{*}{Stratus} & \multirow{2}{*}{\claude{} Sonnet 4.6} & \noiseoff &  5.5:1 & 22.7 \\
                         &                                    & \noiseon &  8.6:1 & 24.2 \\
                         \cmidrule{2-5}
                         & \multirow{2}{*}{\kimi{} K2.5}      & \noiseoff &  8.4:1 & 24.8 \\
                         &                                    & \noiseon &  8.9:1 & 27.6 \\
\midrule
\multirow{2}{*}{Claude Code} & \multirow{2}{*}{\claude{} Sonnet 4.6} & \noiseoff & 13.5:1 & 21.3 \\
                             &                                    & \noiseon & 14.5:1 & 23.9 \\
\midrule
\multirow{2}{*}{Codex} & \multirow{2}{*}{\openai{} GPT-5.4} & \noiseoff &  7.9:1 & 18.9 \\
                       &                                  & \noiseon &  8.1:1 & 21.2 \\
\bottomrule
\end{tabular}
\end{table}

\begin{table} 
\centering
\small
\caption{Top-3 \code{kubectl} read subcommands per agent
    (\% of all read-only \code{kubectl} calls).}
\label{tab:tool-top-reads}
\vspace{3pt}
\begin{tabular}{llcccc}
\toprule
Agent & Model & Noise & 1st & 2nd & 3rd \\
\midrule
\multirow{4}{*}{Stratus} & \multirow{2}{*}{\claude{} Sonnet 4.6} & \noiseoff & get (64) & logs (18) & describe (13) \\
                         &                                    & \noiseon & get (65) & logs (17) & describe (14) \\
                         \cmidrule{2-6}
                         & \multirow{2}{*}{\kimi{} K2.5}      & \noiseoff & get (69) & logs (20) & describe (9) \\
                         &                                    & \noiseon & get (67) & logs (21) & describe (10) \\
\midrule
\multirow{2}{*}{Claude Code} & \multirow{2}{*}{\claude{} Sonnet 4.6} & \noiseoff & get (59) & logs (23) & describe (10) \\
                             &                                    & \noiseon & get (62) & logs (19) & describe (11) \\
\midrule
\multirow{2}{*}{Codex} & \multirow{2}{*}{\openai{} GPT-5.4} & \noiseoff & get (60) & logs (24) & exec (7) \\
                       &                                  & \noiseon & get (57) & logs (25) & exec (8) \\
\bottomrule
\end{tabular}
\end{table}

\begin{table} 
\centering
\small
\caption{Top-5 \code{kubectl} write subcommands per agent
    (\% of all mutating \code{kubectl} calls).}
\label{tab:tool-top-writes}
\vspace{3pt}
\setlength{\tabcolsep}{3pt}
\begin{tabular}{llcccccc}
\toprule
Agent & Model & Noise & 1st & 2nd & 3rd & 4th & 5th \\
\midrule
\multirow{4}{*}{Stratus} & \multirow{2}{*}{\claude{} Sonnet 4.6} & \noiseoff & patch (39) & set (19) & run (14) & delete (9) & rollout (8) \\
                         &                                    & \noiseon & patch (41) & set (23) & rollout (9) & delete (7) & create (6) \\
                         \cmidrule{2-8}
                         & \multirow{2}{*}{\kimi{} K2.5}      & \noiseoff & patch (41) & set (15) & create (13) & rollout (11) & delete (6) \\
                         &                                    & \noiseon & patch (40) & set (13) & delete (11) & create (10) & rollout (9) \\
\midrule
\multirow{2}{*}{Claude Code} & \multirow{2}{*}{\claude{} Sonnet 4.6} & \noiseoff & rollout (35) & patch (32) & run (9) & delete (7) & apply (5) \\
                             &                                    & \noiseon & rollout (40) & patch (29) & delete (11) & run (7) & apply (4) \\
\midrule
\multirow{2}{*}{Codex} & \multirow{2}{*}{\openai{} GPT-5.4} & \noiseoff & rollout (33) & patch (20) & run (19) & port-fwd (12) & delete (5) \\
                       &                                  & \noiseon & rollout (29) & run (21) & patch (17) & port-fwd (11) & delete (11) \\
\bottomrule
\end{tabular}
\end{table}

\begin{table}[H]
\centering
\small
\caption{Mitigation retry analysis (Stratus variants only).
    The total number of mitigation attempts and mitigation reads is 
    conditioned on whether
    the initial diagnosis was correct (D) or
    incorrect ($\neg$ D).}
\label{tab:tool-retries}
\vspace{3pt}
\begin{tabular}{llccccc}
\toprule
 & & & \multicolumn{2}{c}{Attempts} & \multicolumn{2}{c}{Mitig. reads} \\
\cmidrule(lr){4-5} \cmidrule(lr){6-7}
Agent & Model & Noise & D & $\neg$ D & D & $\neg$ D \\
\midrule
\multirow{4}{*}{Stratus} & \multirow{2}{*}{\claude{} Sonnet 4.6} & \noiseoff & 1.88 & 3.82 & 17.3 & 66.2 \\
                         &                                    & \noiseon & 1.55 & 1.61 & 13.2 & 21.6 \\
                         \cmidrule{2-7}
                         & \multirow{2}{*}{\kimi{} K2.5}      & \noiseoff & 1.79 & 1.90 & 19.3 & 26.4 \\
                         &                                    & \noiseon & 1.56 & 1.51 & 16.4 & 21.2 \\
\bottomrule
\end{tabular}
\end{table}

\section{Analysis of Total Tokens and End-to-End Success Rate}
\label{sec:appendix:token-analysis}

\begin{figure}[H]
    \centering
    \includegraphics[width=0.75\linewidth]{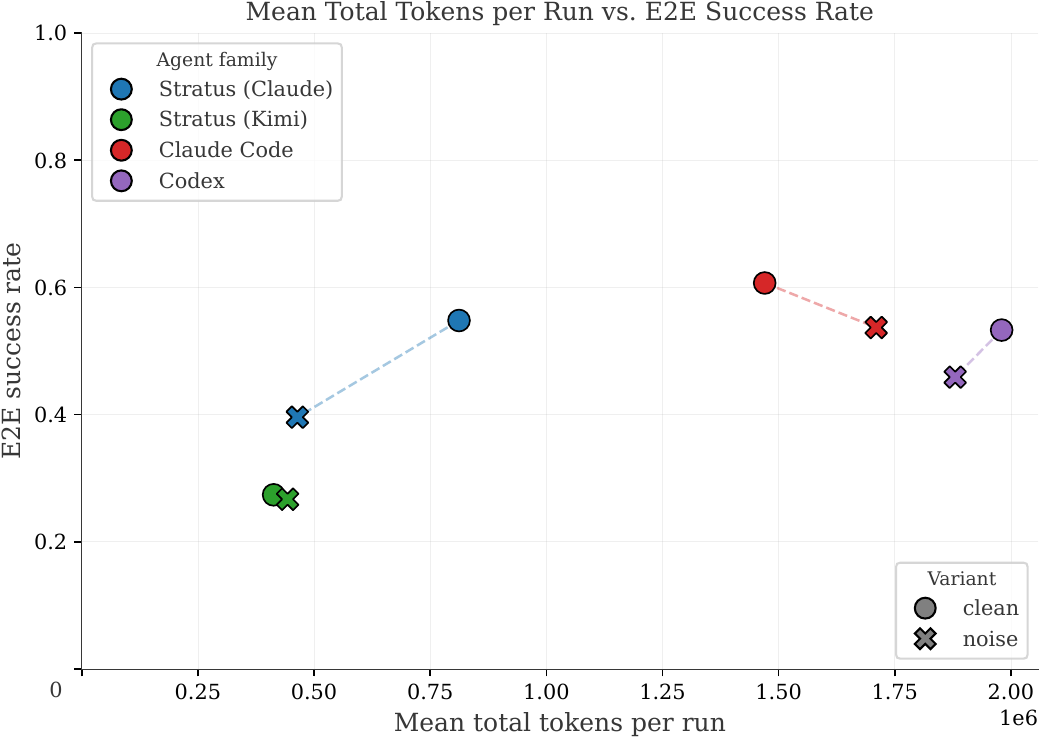}
    \caption{Mean total tokens per run versus end-to-end success
        rate ($P(\text{D} \wedge \text{M})$) for each
        (agent, model, noise) configuration.
        Circles denote the clean condition; crosses denote the
        noisy condition.
        Stratus configurations occupy a low-token / mid-success
        regime, while Claude Code and Codex occupy a high-token
        regime without a corresponding gain in end-to-end success.}
    \label{fig:tokens-vs-e2e}
\end{figure}

Figure~\ref{fig:tokens-vs-e2e} plots mean total tokens per run
    against end-to-end success rate for each
    (agent, model, noise) configuration.
The plot separates two regimes.
Stratus consumes 0.53M tokens per run on average
    (range 0.41--0.81M), while Claude Code uses 1.59M (3.0$\times$)
    and Codex uses 1.93M (3.6$\times$),
    consistent with Stratus's multi-agent architecture that preprocesses
    observability data and prompts the underlying LLM with only the
    filtered subset.
The coding agents instead pull raw observability streams into the
    LLM context, which inflates token usage without a corresponding
    increase in end-to-end success.

Token spend does not predict success across these configurations.
The highest end-to-end rate is achieved by Claude Code in the environment
    with no noise injected, but
    Stratus with Sonnet-4.6 comes within six points of it while
    spending a small fraction of the tokens.
Codex spends the most tokens of any configuration and trails Claude
    Code in end-to-end success, indicating that on \bench{} the marginal
    token does not buy additional success.

Noise reduces the end-to-end success rate for every configuration (the noise marker for each
    color sits below its clean counterpart), and it has varying effects on 
    the token consumption of different agents. 
The coding agents may spend more tokens under noise as
    they ingest additional anomaly signals.
While on paper Stratus with Sonnet-4.6 spends fewer tokens per run under noise, we found that 
    the agent often times out in the 
    noisy condition, indicating that noise substantially increases the 
    amount of investigation Stratus does to diagnose and mitigate the failure.

\section{Broader Impacts}
\label{sec:appendix:broader-impacts}

\bench{} is a benchmark for evaluating AI agents on Site Reliability
    Engineering (SRE). 
We discuss the positive and negative societal impacts that we
    foresee from the work, along with mitigations.

{\bf Positive impacts.}
The most direct positive impact is improving the 
    reliability of AI-assisted computing system operations.
Production failures are a leading cause of 
    service outages,
    financial loss, and engineer/operator burnout. 
A growing number of
    AI products are being marketed for autonomous SRE~\cite{ciroos,devin-oncall,azure-sre-agent,aws-sre-agent}.
Without realistic, mitigation-oriented benchmarks, 
    claims about
    these technologies and products are difficult to verify.
\bench{} is designed to evaluate AI agents'
    capability for autonomous SRE.
It injects faults across the full system stack
    (application, platform, OS, hardware),
    composes them with concurrent noise to model 
    real-world production
    environments, and verifies mitigation against 
    system state.
By making it easier to evaluate SRE agents
    on realistic, high-fidelity SRE problems, we aim 
    to raise
    the empirical bar that AI/agentic SRE products must clear,
    and to reduce the chance that under-tested agents are
    granted production access.
A secondary positive impact is an open infrastructure for research.
The composable APIs, fault and noise injectors,
    and MCP-based agent interface are released so that other
    researchers can extend \bench{} to new applications,
    new fault classes, and new agent architectures without
    rebuilding the scaffolding.
In fact, we have heard from a number of colleagues who are 
    using \bench{} in their research projects.
\bench{} has also been adopted in industry:
    the releases of Devin Oncall~\cite{devin-oncall-blog}
    and Datadog's Scout~1.0~\cite{datadog-scout}
    both report evaluation results on \bench{}.

{\bf Negative impacts and mitigations.}
We considered three categories of potential negative impact.
First, automation displacement: capable AI SRE agents
    could reduce demand for entry-level operations work,
    much like AI coding assistants are reshaping software
    development jobs.
Our results, however, suggest this risk is currently distant:
    even frontier agents struggle on \bench{}'s realistic
    failure scenarios (see \S\ref{sec:eval-setup} and
    Appendix~\ref{sec:case}).
Second, misuse of fault-injection mechanisms:
    \bench{}'s injectors could in principle be repurposed to
    cause harm in production systems that the user controls.
We note that all of the injection primitives we use
    are standard, widely available tools;
    therefore, \bench{} does not introduce a
    novel attack capability beyond what these tools already expose.
The benchmark requires legitimate cluster credentials to
    operate, and mounting a real attack with these primitives
    is no easier than using the underlying tools directly.
Third, over-reliance on benchmark numbers:
    a high \bench{} score is necessary but not sufficient evidence
    that an agent is safe to grant production access.
Appendix~\ref{sec:limitations} explicitly enumerates the assumptions
    and scope under which our scores should be interpreted,
    including applications smaller than production scale,
    a simplified noise model, and a small set of evaluated
    agent-model pairs.
We encourage our users to read \bench{} results as
    a lower bound on the failure modes an agent can recover
    from, not an upper bound on the failure modes it will
    encounter in real-world production deployments.

\section{Limitations}
\label{sec:limitations}

\bench{} has limitations that shape how its
    results should be interpreted and extended.

{\bf Oracle Variance.}
The diagnosis oracle relies on an LLM evaluator,
    which introduces a source of variance that purely
    programmatic checks do not have.
We mitigate this by constraining the LLM evaluator to a structured
    rubric with per-question answers
    (Appendix~\ref{sec:appendix:checklist})
    and by reporting per-dimension breakdowns so that
    readers can see where judgments are stable versus noisy.
Despite its strong empirical reliability (see Table~\ref{tab:judge-reliability}),
    the LLM evaluator remains a best-available approximation for
    evaluating natural-language diagnosis results at scale.

{\bf Noise Modeling.}
The noise injected by \bench{} is a simplification
    of real-world production disturbances.
We inject transient pod crashes and
    resource stress on schedules, which captures one common
    pattern (routine pod churn in a busy system)
    but does not model all sources of production noise,
    such as high-variance traffic anomalies~\cite{meza:osdi:23}, 
    partial network partitions~\cite{Alquraan:osdi:18}, or slow performance degradation
    caused by gradual resource exhaustion~\cite{servicelab}.
Agents that perform well on \bench{}'s noise model may
    still be surprised by noise patterns we do not yet cover.

{\bf System Scale.}
\bench{}'s deployed applications are substantially smaller
    than production systems at major cloud providers.
Our largest application, Train Ticket~\cite{trainticket},
    has 40 microservices,
    whereas production deployments at companies such as
    Uber, Netflix, and Meta commonly run thousands of
    interacting services~\cite{meza:osdi:23,gan:asplos:19}.
Scale affects both the diagnosis 
    (agents face fewer candidate services to investigate)
    and the mitigation evaluation
    (fewer cross-service dependencies to reason about).
Results on \bench{} may not scale linearly to systems that are orders of
    magnitude larger.

{\bf Environment Scope.}
\bench{} targets cloud-native, Kubernetes-based deployments,
    which are the dominant platform for modern production systems
    but not the only one.
Workloads running on monolithic deployments
    or edge deployments have different failure modes
    that \bench{} does not currently exercise.
Extending to these environments would require new fault
    injectors and observability integrations,
    which the composable architecture is designed to support.

\para{Agent Coverage in the Evaluation.}
Constrained by our budget, we cover only four agent-model pairs, drawn from three agents
    (Stratus, Claude Code, Codex)
    and three frontier models
    (Claude Sonnet 4.6, Kimi K2.5, GPT-5.4) in our evaluation.
This is a small sample relative to the space of
    possible agent architectures and LLM backbones,
    and our conclusions about general-purpose coding agents
    versus specialized SRE agents are correspondingly tentative.
We release the benchmark and scoring pipeline so that
    the community can evaluate additional
    agent-model combinations and report results
    under the same oracles.
We are committed to supporting such efforts and maintaining the leaderboard.

\section{Per-Problem End-to-End Results}
\label{sec:appendix:heatmaps}

\begin{figure}
    \centering
    \includegraphics[width=\textwidth]{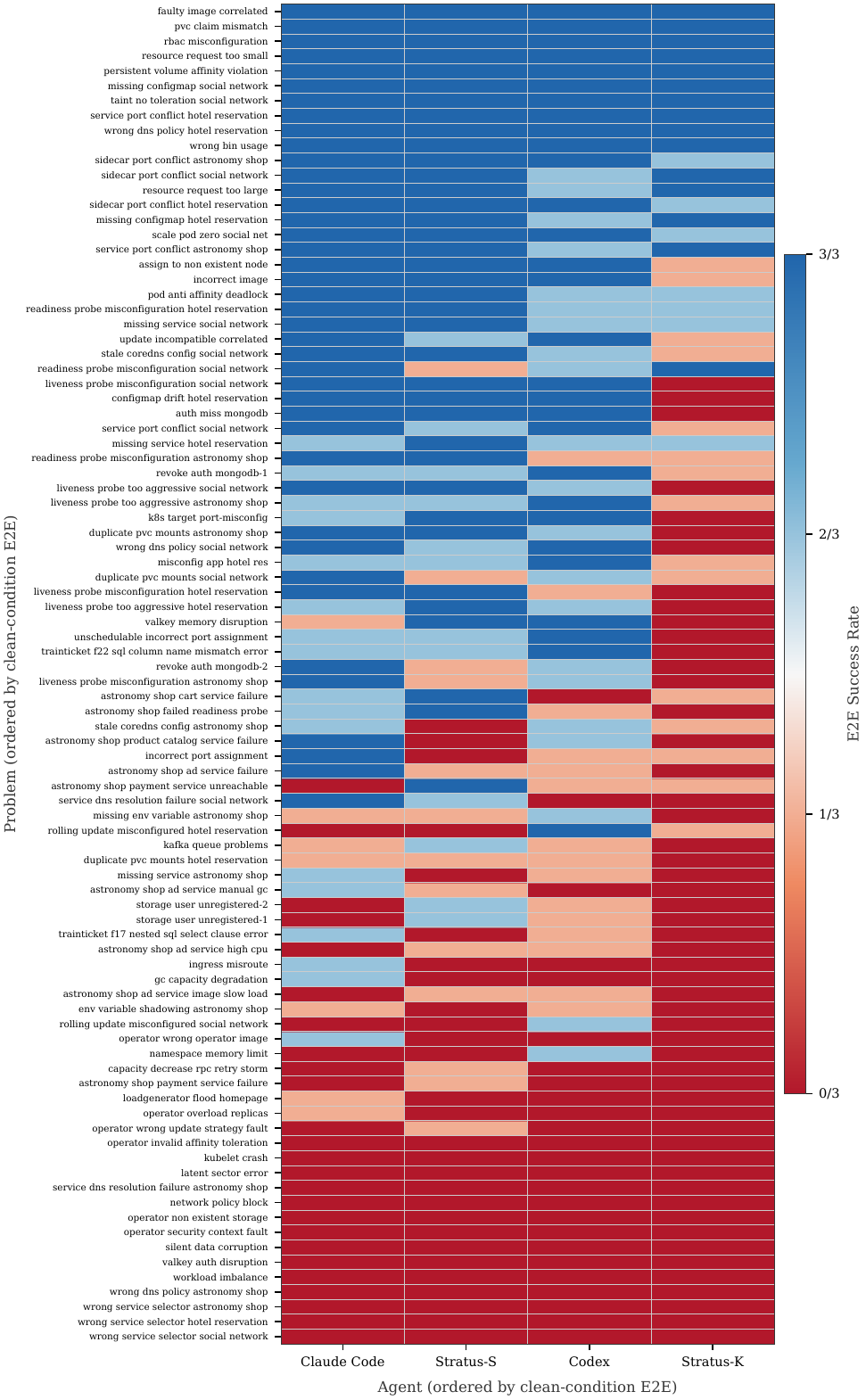}
    \caption{Per-problem end-to-end success rate with no noise injected.}
    \label{fig:heatmap-quiet}
\end{figure}

\begin{figure}
    \centering
    \includegraphics[width=\textwidth]{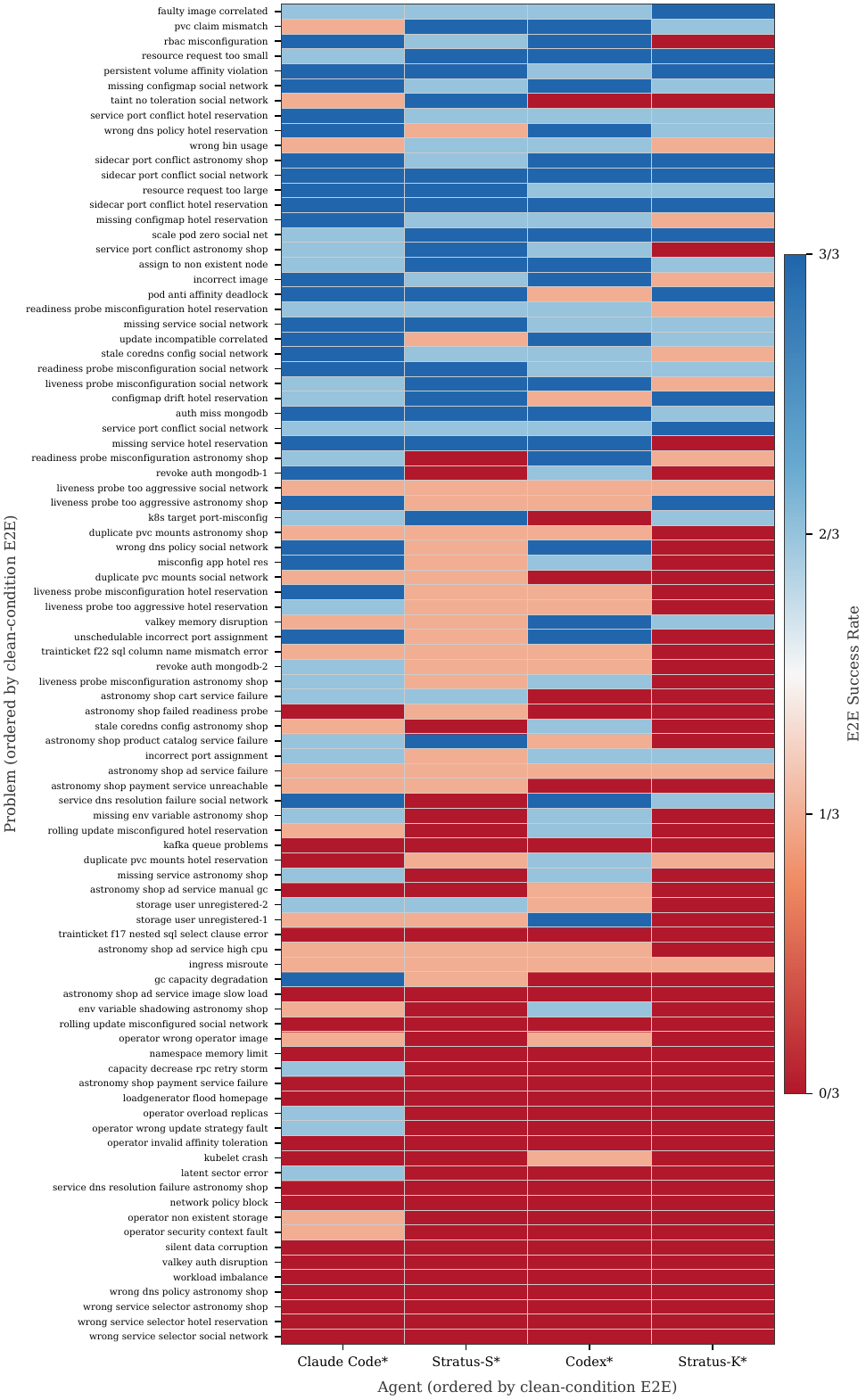}
    \caption{Per-problem end-to-end success rate under noisy conditions.}
    \label{fig:heatmap-noisy}
\end{figure}

Figures~\ref{fig:heatmap-quiet} and~\ref{fig:heatmap-noisy} show the
    per-problem end-to-end (E2E) success rate for each agent
    without and with noise injected into the environment, respectively.
    
A cell value of $k/3$ indicates that $k$ out of three runs achieved
both correct diagnosis and successful mitigation.

\newpage
\section*{NeurIPS Paper Checklist}

\begin{enumerate}

\item {\bf Claims}
    \item[] Question: Do the main claims made in the abstract and introduction accurately reflect the paper's contributions and scope?
    \item[] Answer: \answerYes{}
    \item[] Justification: The abstract and introduction state that \bench{} is a live SRE benchmark with high-fidelity failure scenarios, cross-stack faults, noisy environments, and a composable architecture, and that we evaluate frontier agents on it. These claims map directly to the contributions delivered in \S\ref{sec:platform}--\S\ref{sec:eval-setup}.
    \item[] Guidelines:
    \begin{itemize}
        \item The answer \answerNA{} means that the abstract and introduction do not include the claims made in the paper.
        \item The abstract and/or introduction should clearly state the claims made, including the contributions made in the paper and important assumptions and limitations. A \answerNo{} or \answerNA{} answer to this question will not be perceived well by the reviewers.
        \item The claims made should match theoretical and experimental results, and reflect how much the results can be expected to generalize to other settings.
        \item It is fine to include aspirational goals as motivation as long as it is clear that these goals are not attained by the paper.
    \end{itemize}

\item {\bf Limitations}
    \item[] Question: Does the paper discuss the limitations of the work performed by the authors?
    \item[] Answer: \answerYes{}
    \item[] Justification: Appendix~\ref{sec:limitations} is a dedicated Limitations section covering oracle variance, the simplified noise model, system scale relative to production, scope of supported environments, and the small set of evaluated agents.
    \item[] Guidelines:
    \begin{itemize}
        \item The answer \answerNA{} means that the paper has no limitation while the answer \answerNo{} means that the paper has limitations, but those are not discussed in the paper.
        \item The authors are encouraged to create a separate ``Limitations'' section in their paper.
        \item The paper should point out any strong assumptions and how robust the results are to violations of these assumptions (e.g., independence assumptions, noiseless settings, model well-specification, asymptotic approximations only holding locally). The authors should reflect on how these assumptions might be violated in practice and what the implications would be.
        \item The authors should reflect on the scope of the claims made, e.g., if the approach was only tested on a few datasets or with a few runs. In general, empirical results often depend on implicit assumptions, which should be articulated.
        \item The authors should reflect on the factors that influence the performance of the approach. For example, a facial recognition algorithm may perform poorly when image resolution is low or images are taken in low lighting. Or a speech-to-text system might not be used reliably to provide closed captions for online lectures because it fails to handle technical jargon.
        \item The authors should discuss the computational efficiency of the proposed algorithms and how they scale with dataset size.
        \item If applicable, the authors should discuss possible limitations of their approach to address problems of privacy and fairness.
        \item While the authors might fear that complete honesty about limitations might be used by reviewers as grounds for rejection, a worse outcome might be that reviewers discover limitations that aren't acknowledged in the paper. The authors should use their best judgment and recognize that individual actions in favor of transparency play an important role in developing norms that preserve the integrity of the community. Reviewers will be specifically instructed to not penalize honesty concerning limitations.
    \end{itemize}

\item {\bf Theory assumptions and proofs}
    \item[] Question: For each theoretical result, does the paper provide the full set of assumptions and a complete (and correct) proof?
    \item[] Answer: \answerNA{}
    \item[] Justification: The paper does not contain theoretical results; \bench{} is an empirical benchmark contribution.
    \item[] Guidelines:
    \begin{itemize}
        \item The answer \answerNA{} means that the paper does not include theoretical results.
        \item All the theorems, formulas, and proofs in the paper should be numbered and cross-referenced.
        \item All assumptions should be clearly stated or referenced in the statement of any theorems.
        \item The proofs can either appear in the main paper or the supplemental material, but if they appear in the supplemental material, the authors are encouraged to provide a short proof sketch to provide intuition.
        \item Inversely, any informal proof provided in the core of the paper should be complemented by formal proofs provided in appendix or supplemental material.
        \item Theorems and Lemmas that the proof relies upon should be properly referenced.
    \end{itemize}

    \item {\bf Experimental result reproducibility}
    \item[] Question: Does the paper fully disclose all the information needed to reproduce the main experimental results of the paper to the extent that it affects the main claims and/or conclusions of the paper (regardless of whether the code and data are provided or not)?
    \item[] Answer: \answerYes{}
    \item[] Justification: \S\ref{sec:platform}--\S\ref{sec:implementation} describe the cluster environment, fault and noise injectors, agent interface, and evaluation oracles; \S\ref{sec:eval-setup} specifies agents, models, and run counts; \S\ref{sec:oracle} and Appendix~\ref{sec:appendix:checklist} give the full diagnosis judge protocol and checklist; the per-problem results in Appendix~\ref{sec:appendix:heatmaps} report the underlying numbers. We additionally release the code and configuration as an open-source artifact.
    \item[] Guidelines:
    \begin{itemize}
        \item The answer \answerNA{} means that the paper does not include experiments.
        \item If the paper includes experiments, a \answerNo{} answer to this question will not be perceived well by the reviewers: Making the paper reproducible is important, regardless of whether the code and data are provided or not.
        \item If the contribution is a dataset and\slash or model, the authors should describe the steps taken to make their results reproducible or verifiable.
        \item Depending on the contribution, reproducibility can be accomplished in various ways. For example, if the contribution is a novel architecture, describing the architecture fully might suffice, or if the contribution is a specific model and empirical evaluation, it may be necessary to either make it possible for others to replicate the model with the same dataset, or provide access to the model. In general. releasing code and data is often one good way to accomplish this, but reproducibility can also be provided via detailed instructions for how to replicate the results, access to a hosted model (e.g., in the case of a large language model), releasing of a model checkpoint, or other means that are appropriate to the research performed.
        \item While NeurIPS does not require releasing code, the conference does require all submissions to provide some reasonable avenue for reproducibility, which may depend on the nature of the contribution. For example
        \begin{enumerate}
            \item If the contribution is primarily a new algorithm, the paper should make it clear how to reproduce that algorithm.
            \item If the contribution is primarily a new model architecture, the paper should describe the architecture clearly and fully.
            \item If the contribution is a new model (e.g., a large language model), then there should either be a way to access this model for reproducing the results or a way to reproduce the model (e.g., with an open-source dataset or instructions for how to construct the dataset).
            \item We recognize that reproducibility may be tricky in some cases, in which case authors are welcome to describe the particular way they provide for reproducibility. In the case of closed-source models, it may be that access to the model is limited in some way (e.g., to registered users), but it should be possible for other researchers to have some path to reproducing or verifying the results.
        \end{enumerate}
    \end{itemize}

\item {\bf Open access to data and code}
    \item[] Question: Does the paper provide open access to the data and code, with sufficient instructions to faithfully reproduce the main experimental results, as described in supplemental material?
    \item[] Answer: \answerYes{}
    \item[] Justification: \bench{}, including its fault and noise injectors, evaluation oracles, MCP-based agent interface, and problem definitions, is released as an open-source artifact (at \url{https://github.com/SREGym/SREGym}). Per-problem success heatmaps are reported in Appendix~\ref{sec:appendix:heatmaps}. The Appendix and \S\ref{sec:implementation} describe the deployment, problem authoring, and evaluation procedure needed to reproduce the reported results.
    \item[] Guidelines:
    \begin{itemize}
        \item The answer \answerNA{} means that paper does not include experiments requiring code.
        \item Please see the NeurIPS code and data submission guidelines (\url{https://neurips.cc/public/guides/CodeSubmissionPolicy}) for more details.
        \item While we encourage the release of code and data, we understand that this might not be possible, so \answerNo{} is an acceptable answer. Papers cannot be rejected simply for not including code, unless this is central to the contribution (e.g., for a new open-source benchmark).
        \item The instructions should contain the exact command and environment needed to run to reproduce the results. See the NeurIPS code and data submission guidelines (\url{https://neurips.cc/public/guides/CodeSubmissionPolicy}) for more details.
        \item The authors should provide instructions on data access and preparation, including how to access the raw data, preprocessed data, intermediate data, and generated data, etc.
        \item The authors should provide scripts to reproduce all experimental results for the new proposed method and baselines. If only a subset of experiments are reproducible, they should state which ones are omitted from the script and why.
        \item At submission time, to preserve anonymity, the authors should release anonymized versions (if applicable).
        \item Providing as much information as possible in supplemental material (appended to the paper) is recommended, but including URLs to data and code is permitted.
    \end{itemize}

\item {\bf Experimental setting/details}
    \item[] Question: Does the paper specify all the training and test details (e.g., data splits, hyperparameters, how they were chosen, type of optimizer) necessary to understand the results?
    \item[] Answer: \answerYes{}
    \item[] Justification: We do not train models; \bench{} evaluates off-the-shelf agents and frontier LLMs. \S\ref{sec:eval-setup} lists the evaluated agent--model pairs, the number of runs per problem, the diagnosis-success threshold, the judge model, and the configurations evaluated (clean and noise-injected). The diagnosis rubric, dimension weights, and pass threshold ($\tau{=}\frac{7}{9}$) are specified in \S\ref{sec:oracle} and Appendix~\ref{sec:appendix:checklist}.
    \item[] Guidelines:
    \begin{itemize}
        \item The answer \answerNA{} means that the paper does not include experiments.
        \item The experimental setting should be presented in the core of the paper to a level of detail that is necessary to appreciate the results and make sense of them.
        \item The full details can be provided either with the code, in appendix, or as supplemental material.
    \end{itemize}

\item {\bf Experiment statistical significance}
    \item[] Question: Does the paper report error bars suitably and correctly defined or other appropriate information about the statistical significance of the experiments?
    \item[] Answer: \answerYes{}
    \item[] Justification: Each agent--model--noise configuration is run three times per problem across 90 problems; aggregate success rates, time-to-diagnose, time-to-mitigate, and token usage are reported in Table~\ref{tab:main-results}, with per-problem results in Appendix~\ref{sec:appendix:heatmaps}.
    \item[] Guidelines:
    \begin{itemize}
        \item The answer \answerNA{} means that the paper does not include experiments.
        \item The authors should answer \answerYes{} if the results are accompanied by error bars, confidence intervals, or statistical significance tests, at least for the experiments that support the main claims of the paper.
        \item The factors of variability that the error bars are capturing should be clearly stated (for example, train/test split, initialization, random drawing of some parameter, or overall run with given experimental conditions).
        \item The method for calculating the error bars should be explained (closed form formula, call to a library function, bootstrap, etc.)
        \item The assumptions made should be given (e.g., Normally distributed errors).
        \item It should be clear whether the error bar is the standard deviation or the standard error of the mean.
        \item It is OK to report 1-sigma error bars, but one should state it. The authors should preferably report a 2-sigma error bar than state that they have a 96\% CI, if the hypothesis of Normality of errors is not verified.
        \item For asymmetric distributions, the authors should be careful not to show in tables or figures symmetric error bars that would yield results that are out of range (e.g., negative error rates).
        \item If error bars are reported in tables or plots, the authors should explain in the text how they were calculated and reference the corresponding figures or tables in the text.
    \end{itemize}

\item {\bf Experiments compute resources}
    \item[] Question: For each experiment, does the paper provide sufficient information on the computer resources (type of compute workers, memory, time of execution) needed to reproduce the experiments?
    \item[] Answer: \answerYes{}
    \item[] Justification: Agents are evaluated through hosted model APIs against a Kubernetes-based cluster. Table~\ref{tab:main-results} reports mean time-to-diagnose, mean time-to-mitigate, and mean token usage per run for each agent--model--noise configuration. \bench{} runs on commodity hardware: it has been deployed across a range of configurations, from a single MacBook with an emulated Kubernetes cluster (via Kind) up to multi-node clusters. The results in this paper were obtained on machines provided by CloudLab.
    \item[] Guidelines:
    \begin{itemize}
        \item The answer \answerNA{} means that the paper does not include experiments.
        \item The paper should indicate the type of compute workers CPU or GPU, internal cluster, or cloud provider, including relevant memory and storage.
        \item The paper should provide the amount of compute required for each of the individual experimental runs as well as estimate the total compute.
        \item The paper should disclose whether the full research project required more compute than the experiments reported in the paper (e.g., preliminary or failed experiments that didn't make it into the paper).
    \end{itemize}

\item {\bf Code of ethics}
    \item[] Question: Does the research conducted in the paper conform, in every respect, with the NeurIPS Code of Ethics \url{https://neurips.cc/public/EthicsGuidelines}?
    \item[] Answer: \answerYes{}
    \item[] Justification: The work uses Kubernetes deployments and publicly available open-source applications, does not involve personal data or human subjects, and conforms to the NeurIPS Code of Ethics.
    \item[] Guidelines:
    \begin{itemize}
        \item The answer \answerNA{} means that the authors have not reviewed the NeurIPS Code of Ethics.
        \item If the authors answer \answerNo, they should explain the special circumstances that require a deviation from the Code of Ethics.
        \item The authors should make sure to preserve anonymity (e.g., if there is a special consideration due to laws or regulations in their jurisdiction).
    \end{itemize}

\item {\bf Broader impacts}
    \item[] Question: Does the paper discuss both potential positive societal impacts and negative societal impacts of the work performed?
    \item[] Answer: \answerYes{}
    \item[] Justification: Appendix~\ref{sec:appendix:broader-impacts} discusses positive impacts (raising the empirical bar for AI SRE agents before production deployment, releasing open infrastructure for the community) and negative impacts (potential automation displacement, possible misuse of fault-injection mechanisms, over-reliance on benchmark numbers), along with mitigations.
    \item[] Guidelines:
    \begin{itemize}
        \item The answer \answerNA{} means that there is no societal impact of the work performed.
        \item If the authors answer \answerNA{} or \answerNo, they should explain why their work has no societal impact or why the paper does not address societal impact.
        \item Examples of negative societal impacts include potential malicious or unintended uses (e.g., disinformation, generating fake profiles, surveillance), fairness considerations (e.g., deployment of technologies that could make decisions that unfairly impact specific groups), privacy considerations, and security considerations.
        \item The conference expects that many papers will be foundational research and not tied to particular applications, let alone deployments. However, if there is a direct path to any negative applications, the authors should point it out. For example, it is legitimate to point out that an improvement in the quality of generative models could be used to generate Deepfakes for disinformation. On the other hand, it is not needed to point out that a generic algorithm for optimizing neural networks could enable people to train models that generate Deepfakes faster.
        \item The authors should consider possible harms that could arise when the technology is being used as intended and functioning correctly, harms that could arise when the technology is being used as intended but gives incorrect results, and harms following from (intentional or unintentional) misuse of the technology.
        \item If there are negative societal impacts, the authors could also discuss possible mitigation strategies (e.g., gated release of models, providing defenses in addition to attacks, mechanisms for monitoring misuse, mechanisms to monitor how a system learns from feedback over time, improving the efficiency and accessibility of ML).
    \end{itemize}

\item {\bf Safeguards}
    \item[] Question: Does the paper describe safeguards that have been put in place for responsible release of data or models that have a high risk for misuse (e.g., pre-trained language models, image generators, or scraped datasets)?
    \item[] Answer: \answerNA{}
    \item[] Justification: \bench{} releases benchmark scaffolding (problem definitions, fault and noise injectors, oracles), not pre-trained models or scraped data. The fault-injection mechanisms exposed are standard cloud-native operations primitives (Kubernetes manipulations, Chaos Mesh, eBPF-based syscall instrumentation) already widely available in the chaos-engineering ecosystem.
    \item[] Guidelines:
    \begin{itemize}
        \item The answer \answerNA{} means that the paper poses no such risks.
        \item Released models that have a high risk for misuse or dual-use should be released with necessary safeguards to allow for controlled use of the model, for example by requiring that users adhere to usage guidelines or restrictions to access the model or implementing safety filters.
        \item Datasets that have been scraped from the Internet could pose safety risks. The authors should describe how they avoided releasing unsafe images.
        \item We recognize that providing effective safeguards is challenging, and many papers do not require this, but we encourage authors to take this into account and make a best faith effort.
    \end{itemize}

\item {\bf Licenses for existing assets}
    \item[] Question: Are the creators or original owners of assets (e.g., code, data, models), used in the paper, properly credited and are the license and terms of use explicitly mentioned and properly respected?
    \item[] Answer: \answerYes{}
    \item[] Justification: All third-party applications (DeathStarBench, Train Ticket, Astronomy Shop / OpenTelemetry Demo), platforms (Kubernetes, Helm), data systems (TiDB, MongoDB, Valkey, Kafka), observability backends (Prometheus, Loki, Jaeger), fault-injection tools (Chaos Mesh), and prior benchmarks (AIOpsLab, ITBench, MicroRemed, Cloud-OpsBench) are cited at first mention. We use these assets in accordance with their open-source licenses and document the specific versions in the released artifact.
    \item[] Guidelines:
    \begin{itemize}
        \item The answer \answerNA{} means that the paper does not use existing assets.
        \item The authors should cite the original paper that produced the code package or dataset.
        \item The authors should state which version of the asset is used and, if possible, include a URL.
        \item The name of the license (e.g., CC-BY 4.0) should be included for each asset.
        \item For scraped data from a particular source (e.g., website), the copyright and terms of service of that source should be provided.
        \item If assets are released, the license, copyright information, and terms of use in the package should be provided. For popular datasets, \url{paperswithcode.com/datasets} has curated licenses for some datasets. Their licensing guide can help determine the license of a dataset.
        \item For existing datasets that are re-packaged, both the original license and the license of the derived asset (if it has changed) should be provided.
        \item If this information is not available online, the authors are encouraged to reach out to the asset's creators.
    \end{itemize}

\item {\bf New assets}
    \item[] Question: Are new assets introduced in the paper well documented and is the documentation provided alongside the assets?
    \item[] Answer: \answerYes{}
    \item[] Justification: \bench{} (the benchmark framework, problem registry, fault and noise injectors, MCP-based agent interface, and diagnosis/mitigation oracles) is the new asset introduced. It is documented in \S\ref{sec:platform}--\S\ref{sec:implementation}, with the diagnosis checklist in Appendix~\ref{sec:appendix:checklist}, the coverage breakdown in Appendix~\ref{sec:appendix:coverage}, and per-problem results in Appendix~\ref{sec:appendix:heatmaps}. The released artifact additionally includes a README, problem-authoring guide, and example problem definitions.
    \item[] Guidelines:
    \begin{itemize}
        \item The answer \answerNA{} means that the paper does not release new assets.
        \item Researchers should communicate the details of the dataset\slash code\slash model as part of their submissions via structured templates. This includes details about training, license, limitations, etc.
        \item The paper should discuss whether and how consent was obtained from people whose asset is used.
        \item At submission time, remember to anonymize your assets (if applicable). You can either create an anonymized URL or include an anonymized zip file.
    \end{itemize}

\item {\bf Crowdsourcing and research with human subjects}
    \item[] Question: For crowdsourcing experiments and research with human subjects, does the paper include the full text of instructions given to participants and screenshots, if applicable, as well as details about compensation (if any)?
    \item[] Answer: \answerNA{}
    \item[] Justification: The paper does not involve crowdsourcing or research with human subjects. The diagnosis-judge agreement study (\S\ref{sec:eval-judge-validation}) was conducted by an internal SRE expert (a co-author) re-scoring agent submissions against the same ground truth used by the judge, not by recruited participants.
    \item[] Guidelines:
    \begin{itemize}
        \item The answer \answerNA{} means that the paper does not involve crowdsourcing nor research with human subjects.
        \item Including this information in the supplemental material is fine, but if the main contribution of the paper involves human subjects, then as much detail as possible should be included in the main paper.
        \item According to the NeurIPS Code of Ethics, workers involved in data collection, curation, or other labor should be paid at least the minimum wage in the country of the data collector.
    \end{itemize}

\item {\bf Institutional review board (IRB) approvals or equivalent for research with human subjects}
    \item[] Question: Does the paper describe potential risks incurred by study participants, whether such risks were disclosed to the subjects, and whether Institutional Review Board (IRB) approvals (or an equivalent approval/review based on the requirements of your country or institution) were obtained?
    \item[] Answer: \answerNA{}
    \item[] Justification: The paper does not involve research with human subjects and therefore did not require IRB review.
    \item[] Guidelines:
    \begin{itemize}
        \item The answer \answerNA{} means that the paper does not involve crowdsourcing nor research with human subjects.
        \item Depending on the country in which research is conducted, IRB approval (or equivalent) may be required for any human subjects research. If you obtained IRB approval, you should clearly state this in the paper.
        \item We recognize that the procedures for this may vary significantly between institutions and locations, and we expect authors to adhere to the NeurIPS Code of Ethics and the guidelines for their institution.
        \item For initial submissions, do not include any information that would break anonymity (if applicable), such as the institution conducting the review.
    \end{itemize}

\item {\bf Declaration of LLM usage}
    \item[] Question: Does the paper describe the usage of LLMs if it is an important, original, or non-standard component of the core methods in this research? Note that if the LLM is used only for writing, editing, or formatting purposes and does \emph{not} impact the core methodology, scientific rigor, or originality of the research, declaration is not required.
    \item[] Answer: \answerYes{}
    \item[] Justification: LLMs are central to the methodology in two ways. First, the agents under evaluation are LLM-powered (Stratus with Claude Sonnet 4.6 and Kimi K2.5; Claude Code with Claude Sonnet 4.6; Codex with GPT-5.4); these are declared in \S\ref{sec:eval-setup}. Second, the diagnosis oracle uses an LLM-as-a-judge protocol (Claude Sonnet 4.6) over a structured rubric, described in detail in \S\ref{sec:oracle} and Appendix~\ref{sec:appendix:checklist}, with a human-agreement validation in \S\ref{sec:eval-judge-validation}.
    \item[] Guidelines:
    \begin{itemize}
        \item The answer \answerNA{} means that the core method development in this research does not involve LLMs as any important, original, or non-standard components.
        \item Please refer to our LLM policy in the NeurIPS handbook for what should or should not be described.
    \end{itemize}

\end{enumerate}

\end{document}